\documentclass[10pt,twocolumn,letterpaper]{article}

\usepackage[pagenumbers]{cvpr} %

\usepackage[T1]{fontenc}

\usepackage{multirow,algorithm,algpseudocode,amsmath,mathtools,graphicx,microtype,caption,bbm,textcomp,scalerel,float}
\usepackage{fontawesome5}

\newcommand{\cmt}[1]{\hfill\textit{#1}}   %

\makeatletter
\renewcommand{\paragraph}{%
    \@startsection{paragraph}{4}%
    {\z@}{-0.5em}{-0.5em}%
    {\normalfont\normalsize\bfseries}%
}
\makeatother

\definecolor{cvprblue}{rgb}{0.21,0.49,0.74}

\usepackage[accsupp]{axessibility}  %
\definecolor{cvprblue}{rgb}{0.21,0.49,0.74}
\usepackage[pagebackref,breaklinks,colorlinks,allcolors=cvprblue]{hyperref}

\newcommand{\maxval}[1]{\textbf{#1}}
\newcommand{\secondval}[1]{\underline{#1}}

\def\paperID{39657} %
\def\confName{CVPR}
\def\confYear{2026}

\title{\textcolor{purple}{PRUE}: A \textcolor{purple}{P}ractical \textcolor{purple}{R}ecipe for Field Bo\textcolor{purple}{u}ndary S\textcolor{purple}{e}gmentation at Scale}

\author{
\textbf{Gedeon Muhawenayo}$^{1*}$\textsuperscript{\faEnvelope}, \textbf{Caleb Robinson}$^{2*}$, \textbf{Subash Khanal}$^{3*}$, \textbf{Zhanpei Fang}$^{4*}$,\\
\textbf{Isaac Corley}$^{5}$, \textbf{Alexander Wollam}$^{3}$, \textbf{Tianyi Gao}$^{3}$, \textbf{Leonard Strnad}$^{5}$,
\textbf{Ryan Avery}$^{5}$,\\ 
\textbf{Lyndon Estes}$^{6}$, \textbf{Ana M. Tárano}$^{1}$, \textbf{Nathan Jacobs}$^{3}$, \textbf{Hannah Kerner}$^{1}$\\
\\
$^{1}$Arizona State University \quad
$^{2}$Microsoft AI for Good \quad
$^{3}$Washington University in St. Louis\\
$^{4}$Oregon State University \quad
$^{5}$Wherobots \quad
$^{6}$Clark University
}

\begin{document}

\twocolumn[{
\maketitle

\begin{center}
\includegraphics[height=4.5cm]{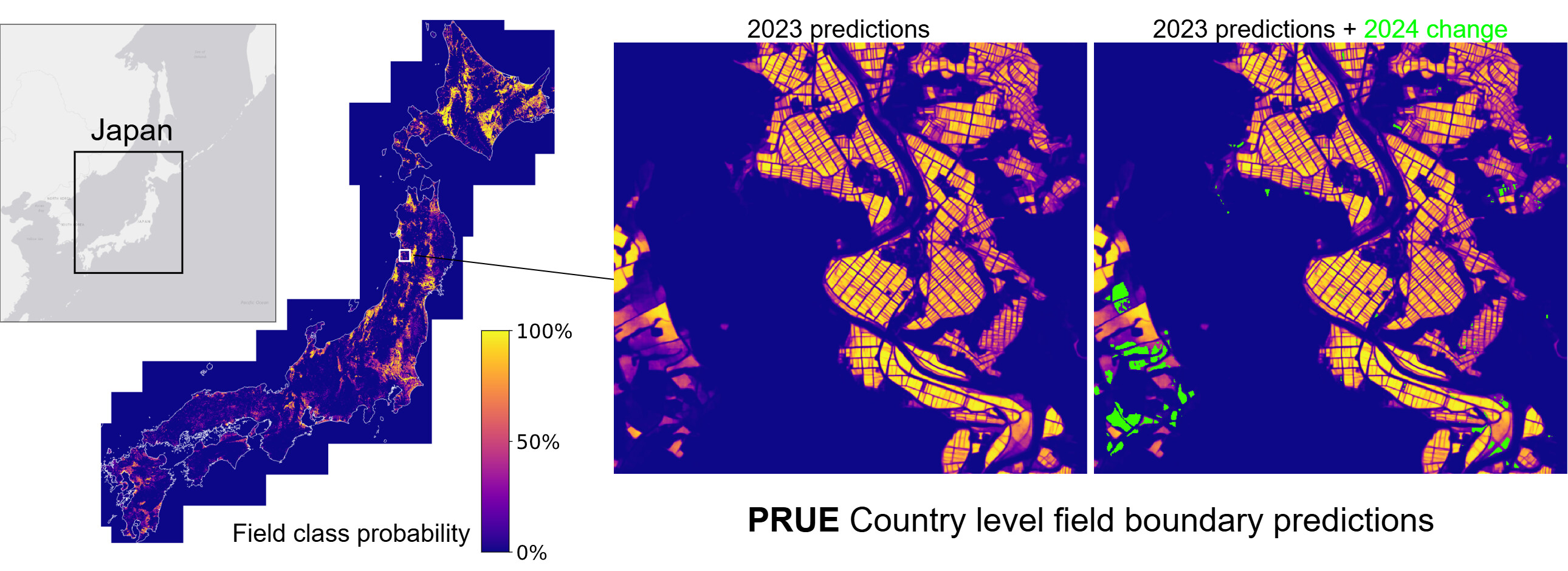}
\end{center}
\captionsetup{type=figure}
\captionof{figure}{%
Example results from the PRUE model in Japan. PRUE specifies a practical recipe for country-scale field boundary segmentation that is robust to reflectance variation and noise in Sentinel-2 L2A imagery, improving standard metrics and large-scale map quality.
}
\label{fig:teaser}

}]

\begin{abstract}
Large-scale maps of field boundaries are essential for agricultural monitoring tasks. Existing deep learning approaches for satellite-based field mapping are sensitive to illumination, spatial scale, and changes in geographic location. We conduct the first systematic evaluation of segmentation and geospatial foundation models (GFMs) for global field boundary delineation using the Fields of The World (FTW) benchmark. We evaluate 18 models under unified experimental settings, showing that a U-Net semantic segmentation model outperforms instance-based and GFM alternatives on a suite of performance and deployment metrics. 
We propose a new segmentation approach that combines a U-Net backbone, composite loss functions, and targeted data augmentations to enhance performance and robustness under real-world conditions. Our model achieves a 76\% IoU and 47\% object-F1 on FTW, an increase of 6\% and 9\% over the previous baseline. 
Our approach provides a practical framework for reliable, scalable, and reproducible field boundary delineation across model design, training, and inference.
We release all models and model-derived field boundary datasets for five countries.
\end{abstract}

\section{Introduction}
\label{sec:intro}
{\def\thefootnote{\faEnvelope}%
\footnotetext{Corresponding Author: gmuhawen@asu.edu}}
{\def\thefootnote{*}%
\footnotetext{Equal Contribution}}

Agricultural field boundary maps (digitized polygons that define individual farm plots) are foundational for agricultural monitoring and decision-making, enabling downstream applications such as  crop type mapping, yield estimation, pest and disease surveillance, and tracking of conservation and climate programs~\cite{Nakalembe_2023}. Field boundary maps enhance national agricultural statistics, providing spatially consistent units for analysis across regions and seasons~\cite{8584043}.

Accurate, up-to-date field boundary maps would transform agriculture and food security applications, but are infeasible with existing methods. Manually delineating field boundaries--typically through image interpretation--is slow, labor-intensive, and must be  regularly repeated as boundaries shift due to land use and management changes~\cite{rs13112197}, making this approach impractical for large-scale monitoring~\cite{7248142}.
Alternatively, machine learning and computer vision methods for satellite imagery enable fast, automated, and repeatable extraction of field boundaries across large regions and over time~\cite{kerner2025fields, essd-15-317-2023}.  Satellite imagery offers wide spatial coverage, frequent revisit intervals, and cost-effective (or free) access to decades of historical observations, making it an ideal data source for scalable agricultural field boundary mapping~\cite{essd-15-317-2023, rs13112197}. 
These advantages have motivated the creation of several open, machine learning-ready field boundary delineation datasets to accelerate progress and benchmarking ~\cite{kerner2025fields, essd-15-317-2023, estes2024regionwidemultiyearsetcrop, 10278130, garnot2021panoptic}. 

However, reliably delineating field boundaries in satellite imagery remains a difficult computer vision problem. Unlike typical instance segmentation tasks, field boundaries are often narrow and poorly defined, varying with crop phenology, management practices, and imaging conditions~\cite{kerner2025fields}. Field edges may appear discontinuous due to mixed pixels, cloud shadows, or low contrast, especially in smallholder systems where fields are heterogeneous and irregularly shaped.  These factors make field delineation a challenging and imperfectly supervised problem, since even expert annotators may disagree on true boundary locations \cite{estes2024regionwidemultiyearsetcrop}.

\begin{figure}[t]
    \centering
    \includegraphics[width=\linewidth]{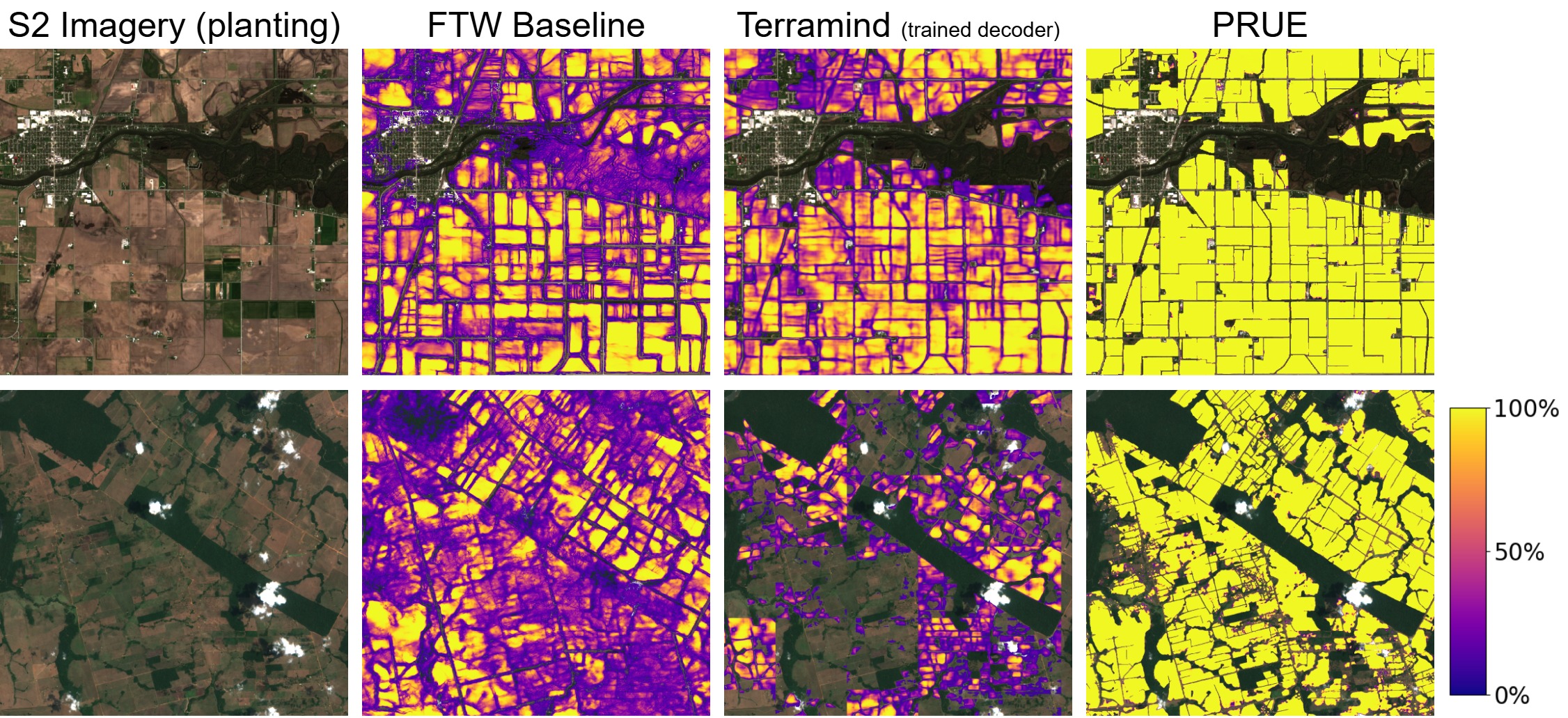}
    \caption{
    Example visualization of predictions over Illinois, USA (top, MGRS tile 16TDL) and Mato Grosso, Brazil (bottom, MGRS tile 20LRQ) with the FTW Baseline, Terramind and PRUE. 
    The FTW Baseline and Terramind models show strong sensitivity to scene characteristics, producing discontinuous and noisy field boundaries. 
    The \textbf{PRUE} model achieves stable boundaries across regions and imaging conditions.}
    \label{fig:artifacts}
\end{figure}

Benchmarks such as Fields of The World (FTW)~\cite{kerner2025fields}, PASTIS~\cite{garnot2021panoptic}, and AI4Boundaries~\cite{essd-15-317-2023} help advance research progress on field boundary segmentation, but do not capture errors that arise when models are deployed for large-scale map-making.  Applying the best FTW model from \citep{kerner2025fields} to map large, out-of-domain regions shows it is sensitive to changes in brightness, temporal ordering, pixel resolution/scale, and receptive field (Figure~\ref{fig:artifacts} shows examples with both FTW Baseline and the Terramind models), leading to patch tiling artifacts and low-quality field boundary maps. Several prior studies have noted similar robustness failures in geospatial deep learning pipelines~\cite{10678215,corley2024changedetectionrealitycheck, rolf2024mission}.

To overcome these obstacles to creating accurate, readily-updated global field boundary datasets with AI, we need robust and scalable models that are: 
\begin{itemize}
    \item Invariant to changes in brightness, spatial scale, seasonal differences, and translations in the input window;
    \item Computationally efficient (low cost) at inference time; 
    \item Performant across diverse agricultural contexts (\eg, very small to very large fields).
\end{itemize}

To develop a model that meets these requirements, we performed a systematic evaluation of diverse segmentation and geospatial foundation model (GFM) architectures for field boundary delineation. 
We conducted experiments that compared an array of GFMs and semantic and instance segmentation architectures, as well as a broad sweep of U-Net model settings using both established performance metrics and a suite of new measures designed to assess the robustness of field segmentation under real-world conditions.
These analyses result in  three key contributions: 
\begin{enumerate}
    \item A new state-of-the-art model for field boundary segmentation (PRUE), which outperforms 18 other models against the FTW benchmark, and has high zero-shot performance. 
    Our model code and weights are available at \href{https://github.com/fieldsoftheworld/ftw-prue}{\textcolor{purple}{github.com/fieldsoftheworld/ftw-prue}}, and a live leaderboard comparing all evaluated models is available at \href{https://huggingface.co/spaces/fieldsoftheworld/ftw-leaderboard}{\textcolor{purple}{huggingface.co/spaces/fieldsoftheworld/ftw-leaderboard}}.
    \item A new set of metrics for evaluating the robustness of field boundary segmentation models to real-world distribution shifts (\eg, changes in brightness or translation).
    \item Country-scale, multi-year field boundary maps for Japan, Mexico, Rwanda, South Africa, and Switzerland, which are publicly available online at \href{https://source.coop/wherobots/fields-of-the-world}{\textcolor{purple}{source.coop/wherobots/fields-of-the-world}}.
    These demonstrate the real-world benefits of PRUE over previous models: the maps predicted using PRUE are more accurate and reveal important landscape changes.
\end{enumerate}

These contributions advance reliable, scalable, and transferable models for delineating field boundaries, enabling the creation of globally consistent, automatically updated datasets that can strengthen equitable and sustainable agricultural decision-making \cite{nakalembe2024}.

\section{Background and Related Work}
\label{sec:related_works}
Early approaches to delineate field boundaries from Earth observations (EO) relied on rule-based image processing, such as edge detection techniques and region-based segmentation~\cite{WATKINS2019294, 964989}. 
Although simple and computationally efficient, these approaches are often ineffective, as their reliance on low-level features makes them sensitive to illumination, noise, and image heterogeneity~\cite{electronics12051156}. These limitations motivated a shift towards data-driven methods that capture spatial context and semantic structure~\cite{electronics12051156, persello2019delineation}.

The success of deep learning for image segmentation has made it the dominant approach for agricultural field delineation~\cite{WALDNER2020111741, persello2019delineation}, which is fundamentally an instance segmentation task focused on identifying individual field polygons. The unique characteristics of satellite imagery have led to diverse methodological approaches, which we organize by their segmentation paradigm.

\paragraph{Semantic segmentation with post-processing.} Many pipelines adopt a semantic segmentation framework, classifying each pixel into two or three classes (\eg, field interior, boundary, background) using models such as random forests \cite{debats2016generalized,estes2022high}, CNNs \cite{persello2019delineation}, and U-Nets \cite{rs14225738, kerner2025fields}. The resulting raster masks are then post-processed with thresholding, connected component analysis, and/or watershed segmentation to recover field instances \cite{rs13112197, 10278130, rs14225738}. Recent refinements incorporate multitask objectives, distance maps, or edge-aware losses that learn contour features, better separate adjacent fields, and provide more accurate polygonization \cite{rs13112197}.
These developments have established U-Net and its variants as the most common field delineation framework  \cite{10278130, rs13112197, rs14225738, kerner2025fields}. 

This predominance of semantic segmentation stems from the challenges of applying standard instance segmentation methods to crop fields, which are often irregularly shaped and densely packed~\cite{Nakalembe_2023, rs13071300, rs14225738}, and thus lack the clear structure that object detection methods exploit through bounding-box regression and non-maximum suppression~\cite{DBLP:journals/corr/abs-2112-11037, schuegraf2022building}. Conversely, semantic outputs may be fragmented, and post-processing introduces hyperparameters (\eg connectivity thresholds) that can affect instance quality.

\paragraph{Instance and panoptic segmentation.}
A smaller body of work applies instance or panoptic segmentation architectures directly, eliminating post-processing steps.
Mask R-CNN has been adapted for field delineation \cite{he2017mask, mei2022using}, but its reliance on bounding boxes and NMS may be suboptimal for irregular geometries.
Recently, Delineate Anything \cite{lavreniuk2025delineateanything} fine-tuned YOLOv11-seg for resolution-agnostic field boundary prediction across multi-sensor European imagery. FieldSeg used Segment Anything (SAM) to delineate Sentinel-2 imagery in 8 study areas across 6 continents \cite{ferreira2025fieldseg}. These studies showed that modern instance segmentation architectures can perform well across diverse landscapes. 
The PASTIS dataset~\cite{garnot2021panoptic} benchmarked panoptic methods on French field parcels from Sentinel-2 time series, while PanopticFPN has been applied to map land cover in aerial imagery~\cite{de2022panoptic}.

\paragraph{Universal segmentation.} 
Task-specific segmentation models lack flexibility to generalize across tasks. 
Universal segmentation models such as Mask2Former \cite{cheng2022masked} and OneFormer \cite{jain2023oneformer} are capable of semantic, instance, \textit{and} panoptic segmentation, but are underexplored for satellite imagery.

\paragraph{Geospatial Foundation Models (GFMs).} GFMs such as SatMAE \cite{satmae2022}, DeCUR \cite{wang2023decur}, Satlas \cite{bastani2023satlaspretrainlargescaledatasetremote}, CROMA \cite{fuller2024croma}, SoftCon \cite{wang2024multilabelguidedsoftcontrastive}, DOFA-v1 \cite{xiong2024neural}, AnySat \cite{astruc2024anysat}, Galileo \cite{tseng2025galileolearningglobal}, Prithvi 2.0 \cite{szwarcman2025prithvieo20versatilemultitemporalfoundation}, Clay \cite{clay2025model}, TerraFM \cite{danish2025terrafmscalablefoundationmodel}, TerraMind \cite{jakubik2025terramindlargescalegenerativemultimodality}, and AlphaEarth Foundations \cite{brown2025alphaearthfoundationsembeddingfield} are pretrained on global-scale EO archives, using contrastive or masked modeling objectives.
Their pretrained encoders provide general-purpose representations that capture land-cover semantics, phenological patterns, and surface-texture variations across geographies. 
GFM embeddings can be leveraged to enhance semantic segmentation performance~\cite{tseng2025galileolearningglobal}, particularly under domain shift or limited labeled data. Previous work has not evaluated GFMs for field boundary segmentation across FTW's diverse regions.

\section{Methods}
\label{sec:methods}
To address the challenges to robust, large-scale inference, we systematically evaluated a range of models against the established FTW benchmark dataset, introducing augmentations and metrics to improve and assess deployment reliability.

\subsection{Dataset}
\label{subsec:dataset}
The Fields of The World (FTW) dataset \cite{kerner2025fields} is a globally distributed benchmark for agricultural field boundary delineation, with over 1.5~million geo-referenced, manually validated field polygons from 24 countries across four continents. 
Each polygon is paired with bi-temporal RGB-NIR (RGBN) Sentinel-2 imagery from the planting and harvest seasons.  
We used FTW's predefined datataset splits, which reduce spatial autocorrelation and enable robust cross-regional evaluation. 
FTW provides a training, validation, and test split for each of the 24 countries in the dataset. \citet{kerner2024accurate} recommends benchmarking models using the average of the individual test accuracies for each country.

\subsection{Model architecture search}
We approached model development as a ``bake-off,'' systematically evaluating an extensive set of semantic segmentation, instance segmentation, and GFM model configurations to find the most practical and robust recipe for field boundary segmentation at scale. We named the winner of the bake-off \textcolor{purple}{PRUE}: A \textcolor{purple}{P}ractical \textcolor{purple}{R}ecipe for Field Bo\textcolor{purple}{u}ndary S\textcolor{purple}{e}gmentation at Scale. For each model configuration, we used the original training settings, including reported hyperparameters and preprocessing, and swept learning rates to ensure fair and reproducible comparison across model families. This approach allowed us to identify the strongest baseline model for subsequent development. 

\paragraph{Semantic segmentation baselines.}
We used the highest-performing model reported by \citet{kerner2025fields} on the FTW benchmark, a U-Net with EfficientNet-B3 backbone, as the ``FTW baseline''. We also evaluated DECODE \cite{rs13112197}, which uses a multitask FracTAL-ResUNet to jointly predict field extent, boundaries, and distance maps. 
As in \citet{kerner2024accurate}, we masked pixels with unknown labels during training for presence-only examples in FTW.
We post-processed all outputs using connected components to extract individual field instances for object-level evaluation (see \S~\ref{subsec:model_metrics}).

\paragraph{Instance and panoptic segmentation baselines.}
We evaluated Delineate Anything \cite{lavreniuk2025delineateanything}, SAM \cite{kirillov2023sam}, and Mask2Former (M2F) \cite{cheng2022masked}. Since Delineate Anything is pretrained for field boundary segmentation and intended to be used in a zero-shot setting, we evaluated it and its smaller variant Del-Any S directly on FTW using the RGB channels from the planting season image. We evaluated SAM in both zero-shot and fine-tuned settings. SAM and M2F (panoptic task) were fine-tuned on the same 8-channel bitemporal input as the U-Net baseline. Due to differences in how instance segmentation models handle training objectives compared to semantic models, we did not mask presence-only examples in training.
See supplement \ref{supp:extended_experiments} for additional results.

\paragraph{Geospatial foundation models (GFMs).}
We computed embeddings from pretrained GFM encoders: Galileo \cite{tseng2025galileolearningglobal}, CROMA \cite{fuller2024croma}, SoftCon \cite{wang2024multilabelguidedsoftcontrastive}, Prithvi 2.0 \cite{szwarcman2025prithvieo20versatilemultitemporalfoundation}, DOFA-v1 \cite{xiong2024neural}, DeCUR \cite{wang2023decur}, Satlas \cite{bastani2023satlaspretrainlargescaledatasetremote}, Clay \cite{clay2025model}, DINOv3 \cite{simeoni2025dinov3}, TerraFM \cite{danish2025terrafmscalablefoundationmodel}, and TerraMind \cite{jakubik2025terramindlargescalegenerativemultimodality}. We obtained token features from pretrained, frozen GFMs for each temporal window. We fused the two windows by concatenating tokens along the feature dimension, and passed the fused tokens through a 3-layer MLP, allowing the model to integrate information from both frames at the token level before decoding. To overcome poor segmentation performance resulting from a 1×1 convolution followed by bilinear upsampling (see Supplemental \ref{supp:extended_ablation}), we adopted a decoder composed of a 3×3 convolutional projection layer, two residual refinement blocks, and a multi-scale convolutional module that expands spatial context, followed by pixel-shuffle upsampling \cite{shi2016real} to produce dense segmentation masks. We used this MLP-based fusion combined with the enhanced convolutional decoder for all GFM experiments with results reported in Table \ref{tab:baselines_results}.

\subsection{Model design space exploration}
\label{subsec:model_design_space_exploration}

Building on the best-performing, most parameter-efficient baseline with high per-km$^2$ throughput, we explored the model design space to identify components critical for accurate, scalable, and robust field boundary delineation. Rather than exhaustively re-tuning every architecture, we used this baseline as a controlled reference point, ensuring that observed trends reflect genuine design effects rather than differences in capacity or optimization. 
We systematically varied architectural, data, and optimization factors to isolate their individual impacts on benchmark accuracy and real-world robustness, with each sweep varying a single factor at a time.  %

\paragraph{Architectures and backbones.}
We compared multiple encoder–decoder architectures, including FCN~\cite{long2015fullyconvolutionalnetworkssemantic}, UPerNet~\cite{xiao2018unifiedperceptualparsingscene}, FCSiam~\cite{8451652}, and U-Net~\cite{ronneberger2015unetconvolutionalnetworksbiomedical} variants. 
We evaluated all U-Net variants with EfficientNet backbones (B3-B7)~\cite{tan2019efficientnet} and Mix Vision Transformers (B2-B5)~\cite{xie2021segformer} to evaluate the impact of increasing capacity relative to the FTW baseline (which used an EfficientNet B3 backbone).

\paragraph{Loss functions.}
We evaluated a range of losses commonly used for field boundary and class imbalance segmentation problems, including cross-entropy (CE), Dice, log-cosh Dice, focal, Tversky, Jaccard, and Fractal Tanimoto (FTNMT) loss functions, as well as their weighted variants ~\cite{Jadon_2020, azad2023lossfunctionserasemantic, rs13183707, salehi2017tverskylossfunctionimage}.

\paragraph{Class weights.}
We further examined the impact of class reweighting by varying the boundary class importance factor $\omega$ in steps of 0.05 within the range $[0.60,0.85]$. The normalized class weights for the three output classes (background, interior, and boundary) were defined as $[0.05, 0.95 - \omega, \omega]$, respectively. This sweep allowed us to assess the sensitivity of model training to the relative emphasis placed on thin boundary regions versus field interiors.

\paragraph{Learning rate.}
We swept learning rates logarithmically to identify stable optimization regimes, testing values in $\{10^{-4}, 3\times10^{-4}, 3\times10^{-3}, 10^{-2}, 3\times10^{-2}\}$ (orders of magnitude common for Adam optimizers in segmentation tasks). We trained each under identical conditions to evaluate convergence stability and sensitivity to step size. 

\paragraph{Data augmentations.}
To address the observed sensitivity of models to variations in brightness, spatial scale, and tiling boundaries, we performed a systematic sweep of data augmentations designed to improve robustness along four key dimensions: input order invariance, brightness robustness, scale robustness, and spatial consistency. We applied each augmentation independently or in combination to evaluate its effect on these robustness properties, which were quantified using the deployment-oriented metrics from \S~\ref{subsec:deployment_metrics}.

\subsection{Evaluation metrics for model comparison}
\label{subsec:model_metrics}
Beyond pixel-level metrics of IoU and F1-score, we report object-level precision and recall from polygonized predictions. We averaged each metric over the individual country test sets in FTW. We excluded ``presence-only'' countries from our evaluation, since only recall can be computed for those countries.
To measure computational efficiency, we report the inference throughput (in km$^2$/s) and number of parameters for each model. 

We computed object precision/recall at a 0.5 confidence threshold, which indicates expected performance under default inference settings. To evaluate the precision-recall tradeoff of changing this threshold, we also computed COCO AP$_{0.5}$ and AP$_{0.5:0.95}$. AP$_{0.5}$ integrates precision across all confidence thresholds at IoU=0.5, and AP$_{0.5:0.95}$ averages this across multiple IoU thresholds ($\{0.5, 0.55, $\dots$, 0.95\}$) to evaluate both confidence calibration and localization quality. 
For semantic segmentation models, which do not produce per-instance confidence scores, we use the mean softmax probability within each polygon (from argmaxed predictions) as a proxy.
Instance segmentation models such as Delineate Anything produce confidence scores with each object detection. SAM has a proxy score of predicted IoU, which is the model's own prediction of mask quality. Similarly, models with a panoptic inference head, such as Mask2Former, output a score for each predicted segment.

\subsection{Deployment-oriented model metrics}
\label{subsec:deployment_metrics}

In geospatial machine learning (GeoML), models are typically trained on patch-based datasets and evaluated with the standard computer vision metrics described above (\eg precision, recall, IoU, and average precision) on held-out patches. 
However, at deployment time,  artifacts can arise when these models are used to construct large maps from entire image scenes \cite{Zvonkov_Tseng_Nakalembe_Kerner_2023,huang2018tiling}. Model performance beyond patch-level measures are critical~\cite{rolf2024mission}.
Specifically, models must be robust to tiling artifacts, invariant to input ordering and preprocessing conventions, and stable under moderate changes in spatial scale.

We propose \textit{deployment-oriented} metrics to complement standard performance metrics. These metrics aim to characterize model behavior at deployment time. These metrics can be computed on the predefined dataset splits used during model development, giving practitioners insight into how models are likely to behave when tiled over large scenes at inference time.

\paragraph{Consistency under translations.}
Modern CNNs and ViTs are not translation equivariant in practice~\cite{ding2023revivingshiftequivariancevision, gruver2024liederivativemeasuringlearned, JMLR:v22:21-0019}. 
Small shifts in input can lead to large changes in output due to padding choices, aliasing from strided downsampling, and architectural elements, such as absolute positional encodings~\cite{azulay2019deep,zhang2019making,shifman2024lost}. In patch-based prediction pipelines, this sensitivity manifests as visible grid artifacts when independently processed patches are stitched into a large map~\cite{huang2018tiling}. Prior work recommends several approaches to reduce translation sensitivity, including strategies that  average  overlapping logits, and variants of sliding-window inference with buffered borders that are discarded during stitching~\cite{huang2018tiling}.

We follow work that measures translation robustness via consistency of model outputs under shifts~\cite{zhang2019making,zou2023delving} and adapt it to patch-based geospatial semantic segmentation. Specifically, we extend the idea of \textit{mean Average Semantic Segmentation Consistency} from~\cite{zou2023delving} and compute prediction agreement across four overlapping corner crops of each patch, rather than two global crops, to better mimic the tiling setup used to create maps at inference time.

\begin{figure}[t]
  \centering
  \includegraphics[width=\linewidth]{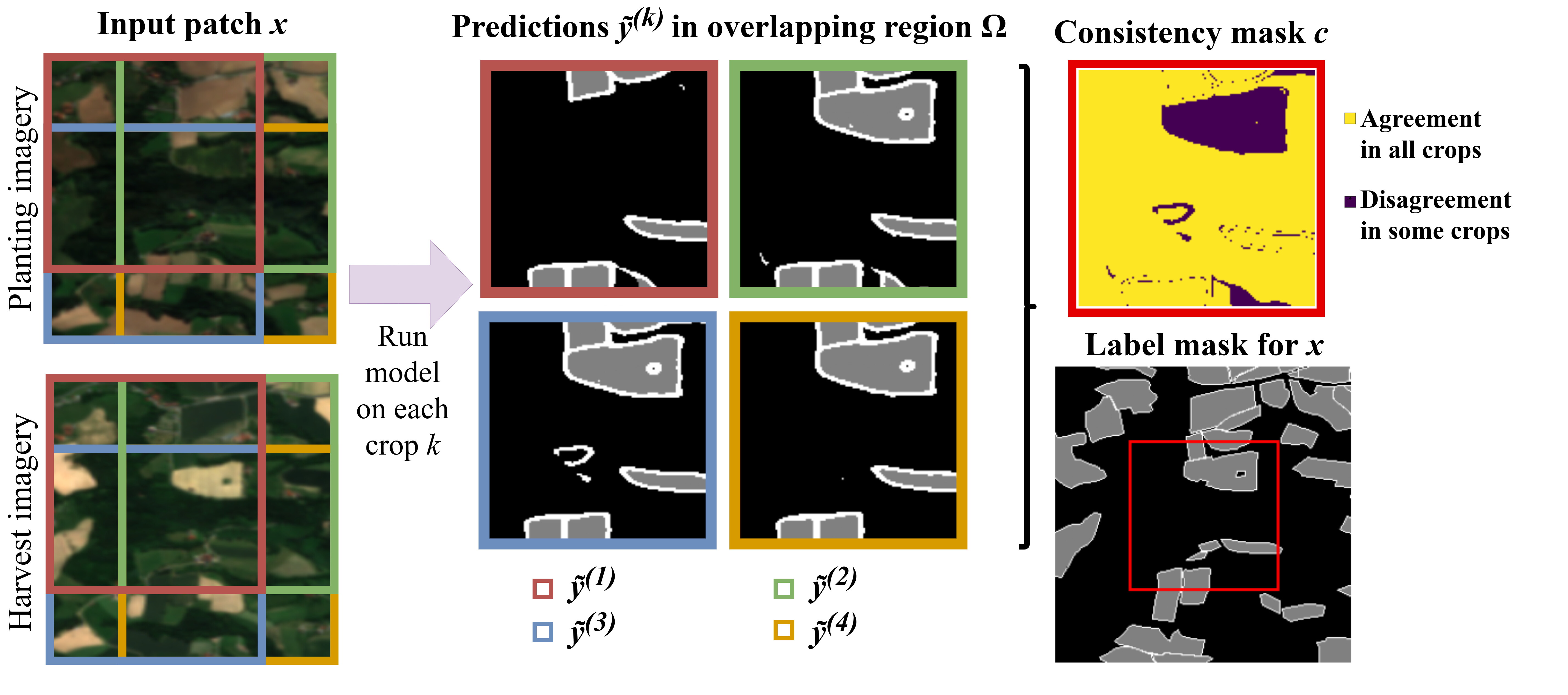}
  \caption{\textbf{Spatial consistency.}
  Four overlapping crops from each image corner are independently segmented. The consistency mask shows pixel-level agreement, with yellow indicating unanimity across all four predictions, and purple disagreement. This metric quantifies grid artifact resistance for large-scale field delineation.}
  \label{fig:consensu_loss}
\end{figure}

Let $x \in \mathbb{R}^{C \times S \times S}$ denote an input patch with height and width $S$. We choose a crop size $p$ such that $S/2 < p < S$ and take four crops of $x$, one anchored at each corner, constructed so that they share a central overlapping region $\Omega$ but differ in their surrounding spatial context. Let $\tilde{y}^{(k)}$ denote the hard per-pixel predictions (\eg, after argmax over logits) obtained from a segmentation model on the $k$-th crop and further cropping the result to the area $\Omega$.
We define the consistency of $x$ as the fraction of pixels in the overlapping region whose predicted labels agree across all four crops:
\begin{equation} \label{eq:consistency}
    \frac{1}{|\Omega|} \sum_{u \in \Omega} 
        \mathbbm{1} \Bigl[
            \tilde{y}^{(1)}(u) = \tilde{y}^{(2)}(u) 
            = \tilde{y}^{(3)}(u) = \tilde{y}^{(4)}(u)
        \Bigr]
\end{equation}
where $u \in \Omega$ indexes over the pixels in the shared area. A consistency of 1 implies the model is perfectly translation-equivariant within the range of shifts implied by the corner crops, while lower values correspond to stronger sensitivity to small translations of the input.
Figure~\ref{fig:consensu_loss} illustrates the cropping scheme and overlapping region $\Omega$ used to compute Eq.~\ref{eq:consistency}.

\paragraph{Sensitivity to input ordering. }Many map prediction problems can be defined to operate on small sets of co-registered observations (\eg, multi-temporal Sentinel-2 scenes or multi-sensor stacks) that are most naturally viewed as unordered sets rather than ordered sequences. In such cases, we would like model predictions to be \textit{permutation invariant} with respect to the ordering of the input elements~\cite{zaheer2017deepsets, segol2020universalequivariantsetnetworks}. 
For example, the FTW dataset~\cite{kerner2025fields} provides paired Sentinel-2 observations from the planting and harvesting stages of the growing season. In the official implementation accompanying the dataset, the input tensor is constructed by stacking the bands in a canonical \textit{(planting, harvest)} ordering~\cite{estes2022high,debats2016generalized}.
At inference time, however, practitioners may select different scenes based on data availability, cloud coverage, and processing level, and may inadvertently permute the temporal order, causing models that are trained in the standard way to fail. For GeoML tasks where inputs consist of observations collected at different times, but preserving temporal ordering is not important, we argue that models should ideally be insensitive to such permutations. Thus, to quantify this property we define an \textit{input-order sensitivity} metric. 

Let $\pi_0$ denote the ``reference'' ordering of input channels (\eg, the training-time convention), and $\Pi$ be a set of alternative permutations of those channels, i.e. the reverse ordering for a pair of observations. For a given evaluation sample $(x,y)$ and performance metric $m$ (\eg, IoU), we compute
\begin{align}
    m_{\text{ref}}(x,y) &= m\bigl(f(x^{\pi_0}),\, y\bigr), \\
    m_{\text{perm}}(x,y)  &= \frac{1}{|\Pi|} \sum_{\pi \in \Pi} m\bigl(f(x^{\pi}),\, y\bigr),
\end{align}
where $x^{\pi}$ denotes the input with channels reordered according to permutation $\pi$. The per-sample order sensitivity is then $\Delta_{\text{order}}(x,y) = m_{\text{ref}}(x,y) - m_{\text{perm}}(x,y)$ and the dataset-level \textit{input-order sensitivity} is the average absolute drop in performance across samples.

\paragraph{Robustness to preprocessing conventions.} 
Satellite imagery is distributed and processed under a variety of radiometric conventions. For example, Sentinel-2 Level-2A products are typically stored as quantized digital numbers with scale factors (\eg, division by $10{,}000$). Starting from Processing Baseline~04.00 (in February 2022), an additive radiometric offset (\eg, \texttt{BOA\_ADD\_OFFSET}) must be applied when converting to physical reflectance~\cite{s2pb04_sentiwiki,s2pb04_forum}. Downstream pipelines may additionally re-scale or normalize the data (\eg, dividing by $3{,}000$ instead of $10{,}000$, or dividing per band by a dataset percentile). 

In practice, training and inference pipelines for GeoML models often do not share identical preprocessing steps, especially when models are reused across organizations, codebases, or data providers. To assess how brittle a model is to such variations, we define a \textit{preprocessing invariance} metric.

Let $g_{\text{ref}}$ denote the ``reference'' normalization used during training (\eg, radiometric offset correction followed by division by $10{,}000$), and let $\{g_j\}_{j=1}^J$ denote $J$ alternative normalizations that reflect plausible deploy-time choices (\eg, different scale factors, omission of offsets, or simple min--max scaling). For each $(x,y)$, we compute
\begin{align}
    m_{\text{ref}}(x,y) &= m\bigl(f(g_{\text{ref}}(x)),\, y\bigr), \\
    m_j(x,y)            &= m\bigl(f(g_j(x)),\, y\bigr) \quad \text{for } j = 1,\dots,J.
\end{align}
We then define the per-sample preprocessing sensitivity as $\Delta_{\text{prep}}(x,y) = \frac{1}{J} \sum_{j=1}^J \bigl| m_{\text{ref}}(x,y) - m_j(x,y) \bigr|$. The dataset-level \textit{preprocessing sensitivity} is the average absolute drop in performance across samples.

\paragraph{Sensitivity to spatial scale.}
GeoML models are frequently trained at a fixed spatial resolution (\eg, $10$\,m Sentinel-2 pixels) but may be deployed on imagery with different effective resolution (\eg, resampled Sentinel-2, or PlanetScope mosaics). Recent self-supervised pretraining approaches for remote sensing explicitly condition ViT models on spatial scale~\cite{reed2023scale}. We measure \textit{scale sensitivity} under test-time resizes in a method similar to the previous metrics. We compute the performance difference between the model run with standard inputs versus resized inputs, $\Delta_{\text{scale}}(x,y)$.
\section{Results and Analysis}
\label{sec:results}

Our results reveal several key insights regarding the performance of different model architectures and design choices (\S\ref{subsec:results-archs-ablations}) based on traditional metrics and our proposed deployment-oriented metrics (\S\ref{subsec:results-final-model}).

\subsection{Model performance \& architecture comparison}
\label{subsec:results-archs-ablations}
Table~\ref{tab:baselines_results} compares our model against semantic, instance, and GFM model baselines on the FTW benchmark.

\paragraph{Architecture families reveal task-dependent strengths.} 
Semantic segmentation models, particularly U-Net variants, remain strong performers on pixel-level and object-level metrics, outperforming instance-based and GFM approaches in pixel-level IoU, precision and recall. Instance segmentation models (Delineate Anything, SAM) achieve reasonable performance in specific contexts with zero-shot settings.
M2F shows competitive object-level performance (precision=0.62, F1=0.39) but lower pixel-level predictions (IoU=0.68). Given their slower inference speeds, these models are less practical for operational deployment; we focused on optimizing the U-Net baseline which exhibits both high accuracy metrics and high throughput (623.28 km$^2/$s).

GFMs paired with semantic decoders achieved moderate performance, but generally underperformed specialized architectures, despite having  3-10$\times$ more parameters than the U-Net baseline. Clay (ViT-L) was the best GFM performer (IoU=0.67, F1=0.36), but was still 9\% and 11\% lower than our optimized U-Net (PRUE). This discrepancy is likely due to the lower effective resolution of GFM encoders, which output coarse-scale patch-wise embeddings. 
Extended results from GFM experiments are shared in the Supplement~\ref{supp:extended_experiments}.

\begin{table*}[!ht]
\centering
\caption{Performance comparison across model families on FTW test set (excluding presence-only countries). 
\textbf{Semantic baselines:} Post-processed with connected components, using presence-only label masking in training \cite{kerner2025fields}. 
\textbf{Instance/panoptic models:} Fine-tuned for 8-channel input and FTW-specific classes but without presence-only masking due to architecture constraints.
\textbf{GFM models:} Frozen encoders with our trained convolutional decoder. Total parameter count includes both the frozen encoder (28M–300M) and our learnable components, which consist of a frames-fusion module and a convolutional decoder (an additional 30M–110M parameters, depending on the encoder’s output dimensionality).
\textbf{Bold} indicates best performance, \underline{underline} indicates second-best.
\small{Throughput measured on a V100-32GB GPU with batch size 64.
\textbf{*}Galileo did not fit into VRAM at any batch size.}
} 

\label{tab:baselines_results}
\resizebox{\textwidth}{!}{%
\begin{tabular}{@{}ll@{\hspace{4pt}}ccc@{\hspace{4pt}}ccccc@{\hspace{3pt}}c@{\hspace{3pt}}c@{}}
\toprule
\multirow{2}{*}{\textbf{Model}} & \multirow{2}{*}{\textbf{Backbone}} & \multicolumn{3}{c}{\textbf{Pixel-level}} & \multicolumn{5}{c}{\textbf{Object-level}} & \textbf{\#Params} & \textbf{Throughput} \\
\cmidrule(lr){3-5} \cmidrule(lr){6-10} \cmidrule(lr){11-11} \cmidrule(lr){12-12}
& & IoU $\uparrow$ & Prec $\uparrow$ & Recall $\uparrow$ & Prec $\uparrow$ & Recall $\uparrow$ & F1 $\uparrow$ & AP$_{0.5:0.95}$ $\uparrow$ & AP$_{0.5}$ $\uparrow$ & (M)$\downarrow$ & (km$^2$/s) $\uparrow$ \\
\midrule
\multicolumn{12}{@{}l}{\textit{\textbf{Semantic segmentation baselines}}} \\
FTW-Baseline & U-Net+EfficientNet-B3 & 0.70 & \underline{0.90} & 0.72 & 0.40 & \underline{0.37} & 0.38 & 0.22 & 0.39 & \underline{13.2} & \textbf{623.28} \\
DECODE & FracTALResUNet & \underline{0.71} & 0.83 & \textbf{0.83} & 0.27 & 0.17 & 0.21 & 0.09 & 0.17 & 64.8 & 113.47 \\
\midrule
\multicolumn{12}{@{}l}{\textit{\textbf{Instance \& panoptic segmentation}}} \\
Mask2Former & Swin-S (fine-tuned, 8ch) & 0.68 & 0.88 & 0.75 & \textbf{0.62} & 0.30 & \underline{0.39} & \textbf{0.28} & \textbf{0.44} & 68.8 & 26.66 \\
SAM & ViT-Huge (fine-tuned, 8ch) & 0.45 & 0.73 & 0.54 & 0.56 & 0.34 & 0.37 & 0.21 & 0.19 & 642.7 & 0.17 \\
SAM & ViT-Huge (zero-shot, 3ch) & 0.32 & 0.36 & \underline{0.82} & 0.14 & 0.34 & 0.17 & 0.06 & 0.12 & 641.1 & 0.17 \\
Del-Any & YOLOv11 (zero-shot, 3ch, win A) & 0.37 & 0.53 & 0.56 & 0.25 & 0.05 & 0.09 & 0.05 & 0.10 & 56.9 & 87.32 \\
Del-Any S & YOLOv11 (zero-shot, 3ch, win A) & 0.44 & 0.52 & 0.73 & 0.15 & 0.06  & 0.08 & 0.07 & 0.14 & \textbf{2.6} & \underline{389.24} \\
\midrule
\multicolumn{12}{@{}l}{\textit{\textbf{Geospatial foundation models}}} \\
Clay & ViT-Large & 0.67 & \textbf{0.91} & 0.72 & 0.38 & 0.36 & 0.36 & 0.24 & \underline{0.41} & 363.8 & 10.98 \\
Galileo & ViT-Base & 0.66 & 0.86 & 0.72 & 0.29 & 0.36 & 0.32 & 0.21 & 0.37 & 119.0 & * \\
DINOv3 & ViT-Large & 0.60 & \underline{0.90} & 0.64 & 0.39 & 0.27 & 0.31 & 0.20 & 0.35 & 412.2 & 46.59 \\
TerraMind & ViT-Base & 0.57 & 0.88 & 0.62 & 0.30 & 0.24 & 0.26 & 0.17 & 0.31 & 189.1 & 123.12 \\
Prithvi 2.0 & ViT-Large & 0.56 & 0.88 & 0.60 & 0.29 & 0.23 & 0.25 & 0.16 & 0.30 & 439.5 & 63.38 \\
TerraFM & ViT-Base & 0.57 & 0.86 & 0.62 & 0.29 & 0.23 & 0.25 & 0.16 & 0.29 & 218.6 & 110.68 \\
CROMA & ViT-Base & 0.52 & 0.86 & 0.56 & 0.26 & 0.18 & 0.21 & 0.13 & 0.25 & 137.0 & 133.10 \\
SoftCon & ViT-Small & 0.52 & 0.85 & 0.57 & 0.24 & 0.18 & 0.21 & 0.12 & 0.24 & 101.8 & 234.18 \\
DeCUR & ViT-Small & 0.49 & 0.85 & 0.53 & 0.23 & 0.16 & 0.19 & 0.11 & 0.22 & 120.2 & 271.67 \\
DOFA-v1 & ViT-Large & 0.49 & 0.85 & 0.53 & 0.21 & 0.15 & 0.17 & 0.10 & 0.19 & 446.0 & 64.85 \\
Satlas & Swin-Tiny & 0.45 & 0.79 & 0.50 & 0.13 & 0.11 & 0.12 & 0.07 & 0.14 & 131.7 & 79.85 \\
\midrule
\textbf{PRUE (ours)} & U-Net+EfficientNet-B7 & \textbf{0.76} & 0.89 & \textbf{0.83} & \textbf{0.62} & \textbf{0.40} & \textbf{0.47} & \underline{0.26} & 0.40 & 67.1 & 306.94 \\
\bottomrule
\end{tabular}}
\end{table*}

\paragraph{Systematic optimization matters more than architectural choice.}
The architecture design search detailed in \S\ref{subsec:model_design_space_exploration} shows that increasing encoder depth and choosing an appropriate loss function can substantially improve delineation quality. Log-cosh Dice loss produces smoother optimization and superior boundary completeness compared to other losses, while moderate boundary class weighting ($\omega=0.75$) yields the most balanced precision–recall trade-off. 
Combining these with brightness and resize augmentations produces consistent performance across domains, confirming that robustness emerges from the interaction of architecture, training objective, and data design choices. 

\subsection{Final model selection and robustness evaluation}
\label{subsec:results-final-model}
Our final model, PRUE, integrates the best-performing design choices: U-Net decoder with EfficientNet-B7 encoder, channel shuffling for input-order invariance, brightness and resize augmentations, log-cosh Dice loss, and boundary weighting $\omega=0.75$. Table~\ref{tab:baselines_results} compares our final model against other backbone architectures; PRUE achieves IoU=0.76 and object F1=0.47, representing $+6\%$ and $+9\%$ improvements over the FTW baseline. We selected U-Net over FCSiam due to higher temporal consistency.

\begin{table*}
\centering
\caption{ Ablation results for controlled experiments on FTW test set (excluding presence-only countries) in which each row varies a single design choice (data augmentations, class weighting, encoder, loss function, or architecture). The \textbf{Combination} rows report the best-performing joint configurations for the FCSiam and U-Net models. \textbf{Bold} indicates best performance, \underline{underline} indicates second-best.}
\resizebox{\textwidth}{!}{%
\begin{tabular}{@{}llccccccccc@{}}
\toprule
   \multirow{2}{*}{\textbf{Category}} &
   \multirow{2}{*}{\textbf{Ablation}} &
  \multicolumn{2}{c}{\textbf{Performance}} &
  \multicolumn{2}{c}{\textbf{Input order}} &
  \multicolumn{2}{c}{\textbf{Brightness}} &
  \multicolumn{2}{c}{\textbf{Scale}} &
  \multicolumn{1}{c}{\textbf{Agree.}} \\

  \cmidrule(lr){3-4} 
  \cmidrule(lr){5-6} 
  \cmidrule(lr){7-8} 
  \cmidrule(lr){9-10} 
  \cmidrule(l){11-11} 

 & &
  Object F1 $\uparrow$ &
  Pixel IoU $\uparrow$ &
  F1$|\Delta|\downarrow$ &
  IoU$|\Delta|\downarrow$ &
  F1$|\Delta|\downarrow$ &
  IoU$|\Delta|\downarrow$ &
  F1$|\Delta|\downarrow$ &
  IoU$|\Delta|\downarrow$ &
  Avg$\uparrow$ \\ 
\midrule

 & FTW-Baseline &
  0.39 $\pm$ 0.08 &
  0.68 $\pm$ 0.08 &
  0.07 &
  0.11 &
  0.04 &
  0.05 &
  0.15 &
  0.12 &
  0.93 \\ 
\midrule

Data augs & Brightness+Resize &
  0.38 $\pm$ 0.08 &
  0.66 $\pm$ 0.09 &
  \underline{0.06} &
  0.10 &
  \underline{0.02} &
  \underline{0.03} &
  \textbf{0.00} &
  \textbf{0.01} &
  \textbf{0.95} \\

Data augs & Channel shuffle &
  0.39 $\pm$ 0.07 &
  0.68 $\pm$ 0.09 &
  \textbf{0.00} &
  \textbf{0.00} &
  0.04 &
  0.05 &
  0.17 &
  0.14 &
  \underline{0.94} \\

Class weights & $\omega=0.75$ &
  0.42 $\pm$ 0.06 &
  0.74 $\pm$ 0.07 &
  0.08 &
  0.11 &
  0.07 &
  0.07 &
  0.29 &
  0.15 &
  \textbf{0.95} \\

Encoder & EfficientNet-B7 &
  0.42 $\pm$ 0.07 &
  0.71 $\pm$ 0.08 &
  0.07 &
  \underline{0.09} &
  0.03 &
  0.04 &
  0.20 &
  0.13 &
  \underline{0.94} \\

Loss function & log-cosh Dice &
  \underline{0.44 $\pm$ 0.07} &
  \textbf{0.77 $\pm$ 0.06} &
  0.09 &
  0.13 &
  0.06 &
  0.05 &
  0.36 &
  0.20 &
  \underline{0.94} \\

Architecture & FCSiam &
  0.40 $\pm$ 0.07 &
  0.69 $\pm$ 0.08 &
  \textbf{0.00} &
  \textbf{0.00} &
  0.05 &
  0.06 &
  0.22 &
  0.14 &
  0.92 \\ 
\midrule

\multirow{2}{*}{Combination} & 
FCSiam combo & %
  \underline{0.44 $\pm$ 0.07} &
  0.75 $\pm$ 0.07 &
  \textbf{0.00} &
  \textbf{0.00} &
  0.04 &
  0.05 &
  0.05 &
  \underline{0.02} &
  \underline{0.94} \\

& \textbf{PRUE} (U-Net) &
  \textbf{0.47 $\pm$ 0.07} &
  \underline{0.76 $\pm$ 0.08} &
  \textbf{0.00} &
  \textbf{0.00} &
  \textbf{0.00} &
  \textbf{0.00} &
  \underline{0.01} &
  \textbf{0.01} &
  \textbf{0.95} \\

\bottomrule
\end{tabular}%
}

\label{tab:ablation_results}
\end{table*}

Table~\ref{tab:ablation_results} shows that PRUE is the most robust configuration across all deployment-oriented perturbations (\S~\ref{subsec:deployment_metrics}). Brightness and resize augmentations reduce sensitivity to illumination and scale changes, while channel shuffling eliminates input-order dependency. Compared to the FTW baseline, PRUE demonstrates negligible variance under input order and brightness shifts, and Figure~\ref{fig:artifacts} shows PRUE's improved generalization compared to the baseline model. These results validate that robustness, accuracy, and scalability are not competing objectives but can be co-optimized through deliberate model and data design. Extended per-country results and ablation analyses are in Supplement~\ref{supp:country_eval}.

\section{AI-Derived Field-Boundaries at Scale}
\label{sec:ai_generated_boundaries}

\paragraph{Country-scale field boundaries.}

We used \textbf{PRUE} to generate complete field boundary maps in 2023 and 2024 for five countries: Japan, Mexico, Rwanda, South Africa, and Switzerland, covering over 4.76 million km$^2$. These countries were selected to cover diverse climatic zones, farming practices, field sizes, and agricultural systems, representing realistic global deployment scenarios. Figure~\ref{fig:teaser} shows examples from Japan. The resulting maps show that the model preserves field topology, maintains coherence across tile boundaries, and generalizes without retraining or regional fine-tuning, thus demonstrating both scalability and zero-shot transferability. 
Table~\ref{tab:ai_stats} gives country-level statistics derived from the mapped field boundaries.

\begingroup
\renewcommand{\arraystretch}{0.85}
\begin{table}[t!]
\centering
\caption{\textbf{Agricultural field statistics derived using PRUE}. The table reports total land area, distributed inference cost, field counts, and median field area in hectares for five diverse countries.}
\label{tab:ai_stats}
\resizebox{\columnwidth}{!}{%
\begin{tabular}{lccccc}
\toprule
\textbf{Country} &
  \textbf{\begin{tabular}[c]{@{}c@{}}Area processed\\(million km$^2$) \end{tabular}} &
  \textbf{\begin{tabular}[c]{@{}c@{}}Cost\\ (\$)\end{tabular}} &
  \textbf{Year} &
  \textbf{\begin{tabular}[c]{@{}c@{}}Fields\\ (M)\end{tabular}} &
  \textbf{\begin{tabular}[c]{@{}c@{}}Median\\ area (ha)\end{tabular}} \\
\midrule
\multirow{2}{*}{Rwanda}       & \multirow{2}{*}{0.02} & \multirow{2}{*}{3.23} & 2023 & 0.18 & 0.05 \\
                              &                       &                       & 2024 & 0.26 & 0.06 \\
\midrule
\multirow{2}{*}{Switzerland}  & \multirow{2}{*}{0.09} & \multirow{2}{*}{5.11} & 2023 & 0.36 & 0.32 \\
                              &                       &                       & 2024 & 0.36 & 0.28 \\
\midrule
\multirow{2}{*}{Japan}        & \multirow{2}{*}{0.65} & \multirow{2}{*}{7.59} & 2023 & 1.55 & 0.20 \\
                              &                       &                       & 2024 & 1.61 & 0.19 \\
\midrule
\multirow{2}{*}{South Africa} & \multirow{2}{*}{1.60} & \multirow{2}{*}{8.03} & 2023 & 2.99 & 0.08 \\
                              &                       &                       & 2024 & 2.70 & 0.07 \\
\midrule
\multirow{2}{*}{Mexico}       & \multirow{2}{*}{2.39} & \multirow{2}{*}{8.26} & 2023 & 5.90 & 0.09 \\
                              &                       &                       & 2024 & 6.57 & 0.09 \\
\bottomrule
\end{tabular}%
}
\end{table}
\endgroup

\paragraph{Mosaicking and inference pipeline.}
To enable large-scale mapping, we developed a pipeline for national-scale inference that emphasizes throughput, spatial consistency, cost efficiency, and reproducibility. The first step generates cloud-free, seasonally aligned Sentinel-2 mosaics using a tiling framework and latitude-based planting/harvest season selection algorithms (see Supplement~\ref{supp:mosaicking_large_scale_inference}). Patches of 256 $\times$ 256 are then read and processed by the model with a 25\% overlap, with Gaussian-weighted averaging applied across overlaps with an apodization kernel~\citep{tolan2024very} that places greater weight on predictions near patch centers, which reduces edge artifacts and ensures consistent boundary logits near patch borders. The stitched probability maps are then vectorized in a blockwise manner using 4,096 × 4,096-pixel windows to maintain memory efficiency. Finally, resulting polygons are serialized in the \texttt{fiboa}~\cite{fiboa2025} GeoParquet format, enabling efficient downstream querying, temporal indexing, and large-scale analytics. This pipeline enables production of contiguous, artifact-free field boundary layers at a country scale.

\paragraph{Throughput and cost efficiency.}
Table~\ref{tab:ai_stats} reports the total land area processed and inference cost for each country.
Creating the planting and harvest mosaics for two agricultural seasons (2023/24-2024/25) 
takes 49.3 minutes. Our inference pipeline is optimized to maximize GPU utilization in a cluster pool of up to 256 NVIDIA A10G GPUs (AWS g5.xlarge instances). Processing the 2-year Mexico mosaic resulted in an execution time of only 14.23 min at a cost of \$8.26 (\$1.05 $\times 10^{-6} / \textrm{km}^2$), consisting of 4.8 GPU-Hrs, 77.9 Core-Hrs, and a throughput of 232.4 GB-Hrs. Using the Machine Learning Impact calculator \cite{lacoste2019quantifying}, we estimate this run generated a total emission of 0.25 kgCO$_2$eq, 100\% of which was offset by AWS.

\paragraph{Field boundary change segmentation}
To quantify structural changes in agricultural landscapes, we computed field-level change directly from the model’s multi-year semantic predictions, which consist of  raster logits produced for the field class for each country and year. We computed  the absolute difference between the two, resulting in a change magnitude map.  We then min–max normalized and thresholded the change at 0.5 to obtain a binary change mask. 
Figure~\ref{fig:teaser} shows an example change map for Japan. Supplement~\ref{supp:qualitative_examples} and~\ref{supp:change_detection} provide additional visualizations and change detection examples across five countries.

\section{Conclusions} \label{sec:conclusions}
We presented a systematic study of model architectures, training strategies, and robustness evaluation metrics for large-scale agricultural field boundary delineation.
Our study establishes a new state-of-the-art on the FTW benchmark, introduces deployment-oriented metrics that better reflect real-world behavior, and demonstrates operational viability through country-scale deployments.

\paragraph{Design choices that matter.}
Loss functions, boundary weighting, and targeted augmentations have the strongest impact on accuracy and robustness. Log-cosh Dice and moderate boundary weighting improve boundary completeness and precision--recall balance, while augmentations targeting deployment failures (brightness, scale, channel shuffling) yield measurable gains. Though explored only for U-Net variants, this design-space methodology is architecture-agnostic and applicable to instance, panoptic, and GFM-based models.

\paragraph{GFMs need task-specific adaptation.}
Despite broad EO pretraining, GFMs generally underperform specialized segmentation models due to resolution limits and weaker localization. They require high-resolution decoders and boundary-aware objectives to match task-specific architectures.

\paragraph{Robustness metrics predict real-world performance.}
Standard metrics do not capture translation sensitivity, input-order dependence, or radiometric brittleness. Our deployment-oriented metrics quantify these behaviors and guide targeted improvements. Models optimized with these metrics show reduced sensitivity to brightness, scale, and translation. Country-scale deployments confirm that higher robustness scores correlate with fewer artifacts and more stable performance across millions of km$^2$.

\section*{Acknowledgments}
This project was supported by funding from the Taylor Geospatial Institute. This research utilized the TGI RAILs advanced computing and data resource, supported by the National Science Foundation (award OAC-2232860) and the Taylor Geospatial Institute. ZF was supported by NASA's Land Cover Land-Use Change Program (award \#80NSSC23K0528). We thank Fuxin Li and Jamon Van Den Hoek for their valuable feedback and suggestions.

{
    \small
    \bibliographystyle{ieeenat_fullname}
    \bibliography{main}

\begin{thebibliography}{78}
\providecommand{\natexlab}[1]{#1}
\providecommand{\url}[1]{\texttt{#1}}
\expandafter\ifx\csname urlstyle\endcsname\relax
  \providecommand{\doi}[1]{doi: #1}\else
  \providecommand{\doi}{doi: \begingroup \urlstyle{rm}\Url}\fi

\bibitem[Astruc et~al.(2025)Astruc, Gonthier, Mallet, and Landrieu]{astruc2024anysat}
Guillaume Astruc, Nicolas Gonthier, Cl\'ement Mallet, and Loic Landrieu.
\newblock {AnySat: One Earth Observation Model for Many Resolutions, Scales, and Modalities}.
\newblock In \emph{Proceedings of the IEEE/CVF Conference on Computer Vision and Pattern Recognition (CVPR)}, pages 19530--19540, 2025.

\bibitem[Azad et~al.(2023)Azad, Heidary, Yilmaz, Hüttemann, Karimijafarbigloo, Wu, Schmeink, and Merhof]{azad2023lossfunctionserasemantic}
Reza Azad, Moein Heidary, Kadir Yilmaz, Michael Hüttemann, Sanaz Karimijafarbigloo, Yuli Wu, Anke Schmeink, and Dorit Merhof.
\newblock {Loss Functions in the Era of Semantic Segmentation: A Survey and Outlook}, 2023.

\bibitem[Azulay and Weiss(2019)]{azulay2019deep}
Aharon Azulay and Yair Weiss.
\newblock Why do deep convolutional networks generalize so poorly to small image transformations?
\newblock \emph{Journal of Machine Learning Research}, 20\penalty0 (184):\penalty0 1--25, 2019.

\bibitem[Bastani et~al.(2023)Bastani, Wolters, Gupta, Ferdinando, and Kembhavi]{bastani2023satlaspretrainlargescaledatasetremote}
Favyen Bastani, Piper Wolters, Ritwik Gupta, Joe Ferdinando, and Aniruddha Kembhavi.
\newblock {SatlasPretrain: A Large-Scale Dataset for Remote Sensing Image Understanding}.
\newblock In \emph{Proceedings of the IEEE/CVF International Conference on Computer Vision (ICCV)}, pages 16772--16782, 2023.

\bibitem[Biscione and Bowers(2021)]{JMLR:v22:21-0019}
Valerio Biscione and Jeffrey~S. Bowers.
\newblock Convolutional neural networks are not invariant to translation, but they can learn to be.
\newblock \emph{Journal of Machine Learning Research}, 22\penalty0 (229):\penalty0 1--28, 2021.

\bibitem[Brown et~al.(2025)Brown, Kazmierski, Pasquarella, Rucklidge, Samsikova, Zhang, Shelhamer, Lahera, Wiles, Ilyushchenko, Gorelick, Zhang, Alj, Schechter, Askay, Guinan, Moore, Boukouvalas, and Kohli]{brown2025alphaearthfoundationsembeddingfield}
Christopher~F. Brown, Michal~R. Kazmierski, Valerie~J. Pasquarella, William~J. Rucklidge, Masha Samsikova, Chenhui Zhang, Evan Shelhamer, Estefania Lahera, Olivia Wiles, Simon Ilyushchenko, Noel Gorelick, Lihui~Lydia Zhang, Sophia Alj, Emily Schechter, Sean Askay, Oliver Guinan, Rebecca Moore, Alexis Boukouvalas, and Pushmeet Kohli.
\newblock {AlphaEarth Foundations: An Embedding Field Model for Accurate and Efficient Global Mapping from Sparse Label Data}, 2025.

\bibitem[Caye~Daudt et~al.(2018)Caye~Daudt, Le~Saux, and Boulch]{8451652}
Rodrigo Caye~Daudt, Bertr Le~Saux, and Alexandre Boulch.
\newblock {Fully Convolutional Siamese Networks for Change Detection}.
\newblock In \emph{2018 25th IEEE International Conference on Image Processing (ICIP)}, pages 4063--4067, 2018.

\bibitem[Cheng et~al.(2022)Cheng, Misra, Schwing, Kirillov, and Girdhar]{cheng2022masked}
Bowen Cheng, Ishan Misra, Alexander~G Schwing, Alexander Kirillov, and Rohit Girdhar.
\newblock {Masked-Attention Mask Transformer for Universal Image Segmentation}.
\newblock In \emph{Proceedings of the IEEE/CVF Conference on Computer Vision and Pattern Recognition}, pages 1290--1299, 2022.

\bibitem[{Clay}(2025)]{clay2025model}
{Clay}.
\newblock {The Clay Foundation Model - An open source AI model and interface for Earth}, 2025.
\newblock Accessed: 2025-11-13.

\bibitem[Cong et~al.(2022)Cong, Khanna, Meng, Liu, Rozi, He, Burke, Lobell, and Ermon]{satmae2022}
Yezhen Cong, Samar Khanna, Chenlin Meng, Patrick Liu, Erik Rozi, Yutong He, Marshall Burke, David~B. Lobell, and Stefano Ermon.
\newblock Sat{MAE}: Pre-training transformers for temporal and multi-spectral satellite imagery.
\newblock In \emph{Advances in Neural Information Processing Systems}, 2022.

\bibitem[Corley et~al.(2024{\natexlab{a}})Corley, Robinson, Dodhia, Lavista~Ferres, and Najafirad]{10678215}
Isaac Corley, Caleb Robinson, Rahul Dodhia, Juan~M. Lavista~Ferres, and Peyman Najafirad.
\newblock Revisiting pre-trained remote sensing model benchmarks: resizing and normalization matters.
\newblock In \emph{2024 IEEE/CVF Conference on Computer Vision and Pattern Recognition Workshops (CVPRW)}, pages 3162--3172, 2024{\natexlab{a}}.

\bibitem[Corley et~al.(2024{\natexlab{b}})Corley, Robinson, and Ortiz]{corley2024changedetectionrealitycheck}
Isaac Corley, Caleb Robinson, and Anthony Ortiz.
\newblock {A Change Detection Reality Check}.
\newblock In \emph{ICLR 2024 Machine Learning for Remote Sensing (ML4RS) Workshop}, 2024{\natexlab{b}}.

\bibitem[d'Andrimont et~al.(2023)d'Andrimont, Claverie, Kempeneers, Muraro, Yordanov, Peressutti, Bati\v{c}, and Waldner]{essd-15-317-2023}
R. d'Andrimont, M. Claverie, P. Kempeneers, D. Muraro, M. Yordanov, D. Peressutti, M. Bati\v{c}, and F. Waldner.
\newblock {AI4Boundaries: An Open AI-Ready Dataset to Map Field Boundaries with Sentinel-2 and Aerial Photography}.
\newblock \emph{Earth System Science Data}, 15\penalty0 (1):\penalty0 317--329, 2023.

\bibitem[Danish et~al.(2026)Danish, Munir, Shah, Khan, Anwer, Laaksonen, Khan, and Khan]{danish2025terrafmscalablefoundationmodel}
Muhammad~Sohail Danish, Muhammad~Akhtar Munir, Syed Roshaan~Ali Shah, Muhammad~Haris Khan, Rao~Muhammad Anwer, Jorma Laaksonen, Fahad~Shahbaz Khan, and Salman Khan.
\newblock Terra{FM}: A scalable foundation model for unified multisensor earth observation.
\newblock In \emph{The Fourteenth International Conference on Learning Representations}, 2026.

\bibitem[de~Carvalho et~al.(2022)de~Carvalho, de~Carvalho~J{\'u}nior, Silva, de~Albuquerque, Santana, Borges, Gomes, and Guimar{\~a}es]{de2022panoptic}
Osmar Luiz~Ferreira de Carvalho, Osmar~Ab{\'\i}lio de Carvalho~J{\'u}nior, Cristiano Rosa~e Silva, Anesmar~Olino de Albuquerque, Nickolas~Castro Santana, Dibio~Leandro Borges, Roberto Arnaldo~Trancoso Gomes, and Renato~Fontes Guimar{\~a}es.
\newblock {Panoptic Segmentation Meets Remote Sensing}.
\newblock \emph{Remote Sensing}, 14\penalty0 (4):\penalty0 965, 2022.

\bibitem[Debats et~al.(2016)Debats, Luo, Estes, Fuchs, and Caylor]{debats2016generalized}
Stephanie~R Debats, Dee Luo, Lyndon~D Estes, Thomas~J Fuchs, and Kelly~K Caylor.
\newblock {A Generalized Computer Vision Approach to Mapping Crop Fields in Heterogeneous Agricultural Landscapes}.
\newblock \emph{Remote Sensing of Environment}, 179:\penalty0 210--221, 2016.

\bibitem[Diakogiannis et~al.(2021)Diakogiannis, Waldner, and Caccetta]{rs13183707}
Foivos~I. Diakogiannis, François Waldner, and Peter Caccetta.
\newblock {Looking for Change? Roll the Dice and Demand Attention}.
\newblock \emph{Remote Sensing}, 13\penalty0 (18), 2021.

\bibitem[Ding et~al.(2023)Ding, Soselia, Armstrong, Su, and Huang]{ding2023revivingshiftequivariancevision}
Peijian Ding, Davit Soselia, Thomas Armstrong, Jiahao Su, and Furong Huang.
\newblock {Reviving Shift Equivariance in Vision Transformers}, 2023.

\bibitem[Erden et~al.(2015)Erden, Aslan, and {\"O}zcanli]{7248142}
Hakan Erden, Murat Aslan, and Cemre~Bahar {\"O}zcanli.
\newblock To establish a new subsidy system.
\newblock In \emph{2015 Fourth International Conference on Agro-Geoinformatics (Agro-geoinformatics)}, pages 57--60, 2015.

\bibitem[Estes et~al.(2022)Estes, Ye, Song, Luo, Eastman, Meng, Zhang, McRitchie, Debats, Muhando, et~al.]{estes2022high}
Lyndon~D Estes, Su Ye, Lei Song, Boka Luo, J~Ronald Eastman, Zhenhua Meng, Qi Zhang, Dennis McRitchie, Stephanie~R Debats, Justus Muhando, et~al.
\newblock {High Resolution, Annual Maps of Field Boundaries for Smallholder-Dominated Croplands at National Scales}.
\newblock \emph{{Frontiers in Artificial Intelligence}}, 4:\penalty0 744863, 2022.

\bibitem[Estes et~al.(2024)Estes, Wussah, Asipunu, Gathigi, KovaÄiÄ, Muhando, Yeboah, Addai, Akakpo, Allotey, Amkoya, Amponsem, Donkoh, Ha, Heltzel, Juma, Mdawida, Miroyo, Mucha, Mugami, Mwawaza, Nyarko, Oduor, Ohemeng, Segbefia, Tumbula, Wambua, Xeflide, Ye, and Yeboah]{estes2024regionwidemultiyearsetcrop}
L.~D. Estes, A. Wussah, M. Asipunu, M. Gathigi, P. KovaÄiÄ, J. Muhando, B.~V. Yeboah, F.~K. Addai, E.~S. Akakpo, M.~K. Allotey, P. Amkoya, E. Amponsem, K.~D. Donkoh, N. Ha, E. Heltzel, C. Juma, R. Mdawida, A. Miroyo, J. Mucha, J. Mugami, F. Mwawaza, D.~A. Nyarko, P. Oduor, K.~N. Ohemeng, S.~I.~D. Segbefia, T. Tumbula, F. Wambua, G.~H. Xeflide, S. Ye, and F. Yeboah.
\newblock {A Region-Wide, Multi-Year Set of Crop Field Boundary Labels for Africa}, 2024.

\bibitem[{European Space Agency}(2022{\natexlab{a}})]{s2pb04_forum}
{European Space Agency}.
\newblock {Introduction of Additional Radiometric Offset in Processing Baseline 04.00 Products}.
\newblock \url{https://forum.step.esa.int/t/info-introduction-of-additional-radiometric-offset-in-pb04-00-products/35431}, 2022{\natexlab{a}}.
\newblock Accessed 9 November 2025.

\bibitem[{European Space Agency}(2022{\natexlab{b}})]{s2pb04_sentiwiki}
{European Space Agency}.
\newblock {Sentinel-2 Processing Baseline and Product Format}.
\newblock \url{https://sentiwiki.copernicus.eu/web/s2-processing}, 2022{\natexlab{b}}.
\newblock Accessed 9 November 2025.

\bibitem[Ferreira et~al.(2025)Ferreira, Martins, Aires, Wijewardane, Zhang, and Samiappan]{ferreira2025fieldseg}
Lucas~B Ferreira, Vitor~S Martins, Uilson~RV Aires, Nuwan Wijewardane, Xin Zhang, and Sathish Samiappan.
\newblock {FieldSeg: A scalable agricultural field extraction framework based on the Segment Anything Model and 10-m Sentinel-2 imagery}.
\newblock \emph{Computers and Electronics in Agriculture}, 232:\penalty0 110086, 2025.

\bibitem[Fuller et~al.(2024)Fuller, Millard, and Green]{fuller2024croma}
Anthony Fuller, Koreen Millard, and James Green.
\newblock {CROMA: Remote Sensing Representations with Contrastive Radar-Optical Masked Autoencoders}.
\newblock \emph{Advances in Neural Information Processing Systems}, 36, 2024.

\bibitem[Garnot and Landrieu(2021)]{garnot2021panoptic}
Vivien Sainte~Fare Garnot and Loic Landrieu.
\newblock {Panoptic Segmentation of Satellite Image Time Series with Convolutional Temporal Attention Networks}.
\newblock In \emph{{Proceedings of the IEEE/CVF International Conference on Computer Vision}}, pages 4872--4881, 2021.

\bibitem[Gruver et~al.(2023)Gruver, Finzi, Goldblum, and Wilson]{gruver2024liederivativemeasuringlearned}
Nate Gruver, Marc~Anton Finzi, Micah Goldblum, and Andrew~Gordon Wilson.
\newblock {The Lie Derivative for Measuring Learned Equivariance}.
\newblock In \emph{The Eleventh International Conference on Learning Representations}, 2023.

\bibitem[He et~al.(2017)He, Gkioxari, Doll{\'a}r, and Girshick]{he2017mask}
Kaiming He, Georgia Gkioxari, Piotr Doll{\'a}r, and Ross Girshick.
\newblock {Mask R-CNN}.
\newblock In \emph{Proceedings of the IEEE International Conference on Computer Vision}, pages 2961--2969, 2017.

\bibitem[Huang et~al.(2018)Huang, Reichman, Collins, Bradbury, and Malof]{huang2018tiling}
Bohao Huang, Daniel Reichman, Leslie~M Collins, Kyle Bradbury, and Jordan~M Malof.
\newblock Tiling and stitching segmentation output for remote sensing: Basic challenges and recommendations.
\newblock \emph{arXiv preprint arXiv:1805.12219}, 2018.

\bibitem[Jadon(2020)]{Jadon_2020}
Shruti Jadon.
\newblock A survey of loss functions for semantic segmentation.
\newblock In \emph{2020 IEEE Conference on Computational Intelligence in Bioinformatics and Computational Biology (CIBCB)}, page 1–7. IEEE, 2020.

\bibitem[Jain et~al.(2023)Jain, Li, Chiu, Hassani, Orlov, and Shi]{jain2023oneformer}
Jitesh Jain, Jiachen Li, MangTik Chiu, Ali Hassani, Nikita Orlov, and Humphrey Shi.
\newblock {OneFormer: One Transformer to Rule Universal Image Segmentation}.
\newblock In \emph{CVPR}, 2023.

\bibitem[Jakubik et~al.(2025)Jakubik, Yang, Blumenstiel, Scheurer, Sedona, Maurogiovanni, Bosmans, Dionelis, Marsocci, Kopp, et~al.]{jakubik2025terramindlargescalegenerativemultimodality}
Johannes Jakubik, Felix Yang, Benedikt Blumenstiel, Erik Scheurer, Rocco Sedona, Stefano Maurogiovanni, Jente Bosmans, Nikolaos Dionelis, Valerio Marsocci, Niklas Kopp, et~al.
\newblock Terramind: Large-scale generative multimodality for earth observation.
\newblock \emph{IEEE/CVF International Conference on Computer Vision (ICCV)}, 2025.

\bibitem[Kerner et~al.(2024)Kerner, Nakalembe, Yang, Zvonkov, McWeeny, Tseng, and Becker-Reshef]{kerner2024accurate}
Hannah Kerner, Catherine Nakalembe, Adam Yang, Ivan Zvonkov, Ryan McWeeny, Gabriel Tseng, and Inbal Becker-Reshef.
\newblock {How Accurate are Existing Land Cover Maps for Agriculture in Sub-Saharan Africa?}
\newblock \emph{Scientific Data}, 11\penalty0 (1):\penalty0 486, 2024.

\bibitem[Kerner et~al.(2025)Kerner, Chaudhari, Ghosh, Robinson, Ahmad, Choi, Jacobs, Holmes, Mohr, Dodhia, et~al.]{kerner2025fields}
Hannah Kerner, Snehal Chaudhari, Aninda Ghosh, Caleb Robinson, Adeel Ahmad, Eddie Choi, Nathan Jacobs, Chris Holmes, Matthias Mohr, Rahul Dodhia, et~al.
\newblock {Fields of the World: A Machine Learning Benchmark Dataset For Global Agricultural Field Boundary Segmentation}.
\newblock In \emph{Proceedings of the AAAI Conference on Artificial Intelligence}, pages 28151--28159, 2025.

\bibitem[Kirillov et~al.(2023)Kirillov, Mintun, Ravi, Mao, Rolland, Gustafson, Xiao, Whitehead, Berg, Lo, Dollar, and Girshick]{kirillov2023sam}
Alexander Kirillov, Eric Mintun, Nikhila Ravi, Hanzi Mao, Chloe Rolland, Laura Gustafson, Tete Xiao, Spencer Whitehead, Alexander~C. Berg, Wan-Yen Lo, Piotr Dollar, and Ross Girshick.
\newblock {Segment Anything}, 2023.

\bibitem[Lacoste et~al.(2019)Lacoste, Luccioni, Schmidt, and Dandres]{lacoste2019quantifying}
Alexandre Lacoste, Alexandra Luccioni, Victor Schmidt, and Thomas Dandres.
\newblock {Quantifying the Carbon Emissions of Machine Learning}.
\newblock \emph{arXiv preprint arXiv:1910.09700}, 2019.

\bibitem[Lavreniuk et~al.(2025)Lavreniuk, Kussul, Shelestov, Yailymov, Salii, Kuzin, and Szantoi]{lavreniuk2025delineateanything}
Mykola Lavreniuk, Nataliia Kussul, Andrii Shelestov, Bohdan Yailymov, Yevhenii Salii, Volodymyr Kuzin, and Zoltan Szantoi.
\newblock {Delineate Anything: Resolution-Agnostic Field Boundary Delineation on Satellite Imagery}.
\newblock \emph{arXiv preprint arXiv:2504.02534}, 2025.

\bibitem[Long et~al.(2015)Long, Shelhamer, and Darrell]{long2015fullyconvolutionalnetworkssemantic}
Jonathan Long, Evan Shelhamer, and Trevor Darrell.
\newblock Fully convolutional networks for semantic segmentation.
\newblock In \emph{2015 IEEE Conference on Computer Vision and Pattern Recognition (CVPR)}, pages 3431--3440, 2015.

\bibitem[Mei et~al.(2022)Mei, Wang, Fouhey, Zhou, Hinks, Gray, Van~Berkel, and Jain]{mei2022using}
Weiye Mei, Haoyu Wang, David Fouhey, Weiqi Zhou, Isabella Hinks, Josh~M Gray, Derek Van~Berkel, and Meha Jain.
\newblock {Using Deep Learning and Very-High-Resolution Imagery to Map Smallholder Field Boundaries}.
\newblock \emph{Remote Sensing}, 14\penalty0 (13):\penalty0 3046, 2022.

\bibitem[Nakalembe and Kerner(2023)]{Nakalembe_2023}
Catherine Nakalembe and Hannah Kerner.
\newblock {Considerations for AI-EO for Agriculture in Sub-Saharan Africa}.
\newblock \emph{Environmental Research Letters}, 18\penalty0 (4):\penalty0 041002, 2023.

\bibitem[Nakalembe et~al.(2024)Nakalembe, Kerner, Zvonkov, et~al.]{nakalembe2024}
Catherine Nakalembe, Hannah Kerner, Ivan Zvonkov, et~al.
\newblock {A Framework for EO-Based National Agricultural Monitoring (EO-NAM) - For the African Context}, 2024.
\newblock preprint.

\bibitem[North et~al.(2019)North, Pairman, and Belliss]{8584043}
Heather~C. North, David Pairman, and Stella~E. Belliss.
\newblock {Boundary Delineation of Agricultural Fields in Multitemporal Satellite Imagery}.
\newblock \emph{IEEE Journal of Selected Topics in Applied Earth Observations and Remote Sensing}, 12\penalty0 (1):\penalty0 237--251, 2019.

\bibitem[Olofsson et~al.(2014)Olofsson, Foody, Herold, Stehman, Woodcock, and Wulder]{olofsson2014good}
Pontus Olofsson, Giles~M Foody, Martin Herold, Stephen~V Stehman, Curtis~E Woodcock, and Michael~A Wulder.
\newblock Good practices for estimating area and assessing accuracy of land change.
\newblock \emph{Remote Sensing of Environment}, 148:\penalty0 42--57, 2014.

\bibitem[Persello et~al.(2019)Persello, Tolpekin, Bergado, and De~By]{persello2019delineation}
Claudio Persello, Valentyn~A Tolpekin, J~Ray Bergado, and Rolf~A De~By.
\newblock {Delineation of Agricultural Fields in Smallholder Farms from Satellite Images Using Fully Convolutional Networks and Combinatorial Grouping}.
\newblock \emph{Remote Sensing of Environment}, 231:\penalty0 111253, 2019.

\bibitem[Persello et~al.(2023)Persello, Grift, Fan, Paris, H{\"a}nsch, Koeva, and Nelson]{10278130}
Claudio Persello, Jeroen Grift, Xinyan Fan, Claudia Paris, Ronny H{\"a}nsch, Mila Koeva, and Andrew Nelson.
\newblock {AI4SmallFarms: A Dataset for Crop Field Delineation in Southeast Asian Smallholder Farms}.
\newblock \emph{IEEE Geoscience and Remote Sensing Letters}, 20:\penalty0 1--5, 2023.

\bibitem[Radoux and Bogaert(2017)]{radoux2017good}
Julien Radoux and Patrick Bogaert.
\newblock {Good Practices for Object-Based Accuracy Assessment}.
\newblock \emph{Remote Sensing}, 9\penalty0 (7):\penalty0 646, 2017.

\bibitem[Ravi et~al.(2024)Ravi, Gabeur, Hu, Hu, Ryali, Ma, Khedr, Rädle, Rolland, Gustafson, Mintun, Pan, Alwala, Carion, Wu, Girshick, Dollár, and Feichtenhofer]{ravi2024sam2segmentimages}
Nikhila Ravi, Valentin Gabeur, Yuan-Ting Hu, Ronghang Hu, Chaitanya Ryali, Tengyu Ma, Haitham Khedr, Roman Rädle, Chloe Rolland, Laura Gustafson, Eric Mintun, Junting Pan, Kalyan~Vasudev Alwala, Nicolas Carion, Chao-Yuan Wu, Ross Girshick, Piotr Dollár, and Christoph Feichtenhofer.
\newblock Sam 2: Segment anything in images and videos, 2024.

\bibitem[Reed et~al.(2023)Reed, Gupta, Li, Brockman, Funk, Clipp, Keutzer, Candido, Uyttendaele, and Darrell]{reed2023scale}
Colorado~J Reed, Ritwik Gupta, Shufan Li, Sarah Brockman, Christopher Funk, Brian Clipp, Kurt Keutzer, Salvatore Candido, Matt Uyttendaele, and Trevor Darrell.
\newblock {Scale-MAE: A Scale-Aware Masked Autoencoder for Multiscale Geospatial Representation Learning}.
\newblock In \emph{Proceedings of the IEEE/CVF International Conference on Computer Vision}, pages 4088--4099, 2023.

\bibitem[Rolf et~al.(2024)Rolf, Klemmer, Robinson, and Kerner]{rolf2024mission}
Esther Rolf, Konstantin Klemmer, Caleb Robinson, and Hannah Kerner.
\newblock {Position: Mission Critical--Satellite Data is a Distinct Modality in Machine Learning}.
\newblock In \emph{41st International Conference on Machine Learning}, 2024.

\bibitem[Ronneberger et~al.(2015)Ronneberger, Fischer, and Brox]{ronneberger2015unetconvolutionalnetworksbiomedical}
Olaf Ronneberger, Philipp Fischer, and Thomas Brox.
\newblock {U-Net: Convolutional networks for biomedical image segmentation}.
\newblock In \emph{International Conference on Medical image computing and computer-assisted intervention}, pages 234--241. Springer, 2015.

\bibitem[Rydberg and Borgefors(2001)]{964989}
A. Rydberg and G. Borgefors.
\newblock {Integrated Method for Boundary Delineation of Agricultural Fields in Multispectral Satellite Images}.
\newblock \emph{IEEE Transactions on Geoscience and Remote Sensing}, 39\penalty0 (11):\penalty0 2514--2520, 2001.

\bibitem[Salehi et~al.(2017)Salehi, Erdogmus, and Gholipour]{salehi2017tverskylossfunctionimage}
Seyed Sadegh~Mohseni Salehi, Deniz Erdogmus, and Ali Gholipour.
\newblock {Tversky loss function for image segmentation using 3D fully convolutional deep networks}, 2017.

\bibitem[Schuegraf et~al.(2022)Schuegraf, Schnell, Henry, and Bittner]{schuegraf2022building}
Philipp Schuegraf, Julian Schnell, Corentin Henry, and Ksenia Bittner.
\newblock {Building Section Instance Segmentation with Combined Classical and Deep Learning Methods}.
\newblock \emph{ISPRS Annals of the Photogrammetry, Remote Sensing and Spatial Information Sciences}, 2:\penalty0 407--414, 2022.

\bibitem[Segol and Lipman(2020)]{segol2020universalequivariantsetnetworks}
Nimrod Segol and Yaron Lipman.
\newblock On universal equivariant set networks.
\newblock In \emph{International Conference on Learning Representations}, 2020.

\bibitem[Shi et~al.(2016)Shi, Caballero, Husz{\'a}r, Totz, Aitken, Bishop, Rueckert, and Wang]{shi2016real}
Wenzhe Shi, Jose Caballero, Ferenc Husz{\'a}r, Johannes Totz, Andrew~P Aitken, Rob Bishop, Daniel Rueckert, and Zehan Wang.
\newblock {Real-Time Single Image and Video Super-Resolution Using an Efficient Sub-Pixel Convolutional Neural Network}.
\newblock In \emph{Proceedings of the IEEE Conference on Computer Vision and Pattern Recognition}, pages 1874--1883, 2016.

\bibitem[Shifman and Weiss(2024)]{shifman2024lost}
Ofir Shifman and Yair Weiss.
\newblock Lost in translation: Modern neural networks still struggle with small realistic image transformations.
\newblock In \emph{European Conference on Computer Vision}, pages 231--247. Springer, 2024.

\bibitem[Sim{\'e}oni et~al.(2025)Sim{\'e}oni, Vo, Seitzer, Baldassarre, Oquab, Jose, Khalidov, Szafraniec, Yi, Ramamonjisoa, Massa, Haziza, Wehrstedt, Wang, Darcet, Moutakanni, Sentana, Roberts, Vedaldi, Tolan, Brandt, Couprie, Mairal, J{\'e}gou, Labatut, and Bojanowski]{simeoni2025dinov3}
Oriane Sim{\'e}oni, Huy~V. Vo, Maximilian Seitzer, Federico Baldassarre, Maxime Oquab, Cijo Jose, Vasil Khalidov, Marc Szafraniec, Seungeun Yi, Micha{\"e}l Ramamonjisoa, Francisco Massa, Daniel Haziza, Luca Wehrstedt, Jianyuan Wang, Timoth{\'e}e Darcet, Th{\'e}o Moutakanni, Leonel Sentana, Claire Roberts, Andrea Vedaldi, Jamie Tolan, John Brandt, Camille Couprie, Julien Mairal, Herv{\'e} J{\'e}gou, Patrick Labatut, and Piotr Bojanowski.
\newblock {DINOv3}, 2025.

\bibitem[Szwarcman et~al.(2026)Szwarcman, Roy, Fraccaro, Gíslason, Blumenstiel, Ghosal, de~Oliveira, de~Sousa~Almeida, Sedona, Kang, Chakraborty, Wang, Gomes, Kumar, Gaur, Truong, Godwin, Khallaghi, Lee, Hsu, Asanjan, Mujeci, Shidham, Balogun, Kolluru, Keenan, Arevalo, Li, Alemohammad, Olofsson, Mayer, Hain, Kennedy, Zadrozny, Bell, Cavallaro, Watson, Maskey, Ramachandran, and Moreno]{szwarcman2025prithvieo20versatilemultitemporalfoundation}
Daniela Szwarcman, Sujit Roy, Paolo Fraccaro, Þorsteinn~Elí Gíslason, Benedikt Blumenstiel, Rinki Ghosal, Pedro~Henrique de Oliveira, Joao~Lucas de Sousa~Almeida, Rocco Sedona, Yanghui Kang, Srija Chakraborty, Sizhe Wang, Carlos Gomes, Ankur Kumar, Vishal Gaur, Myscon Truong, Denys Godwin, Sam Khallaghi, Hyunho Lee, Chia-Yu Hsu, Ata~Akbari Asanjan, Besart Mujeci, Disha Shidham, Rufai~Omowunmi Balogun, Venkatesh Kolluru, Trevor Keenan, Paulo Arevalo, Wenwen Li, Hamed Alemohammad, Pontus Olofsson, Timothy Mayer, Christopher Hain, Robert Kennedy, Bianca Zadrozny, David Bell, Gabriele Cavallaro, Campbell Watson, Manil Maskey, Rahul Ramachandran, and Juan~Bernabe Moreno.
\newblock Prithvi-eo-2.0: A versatile multitemporal foundation model for earth observation applications.
\newblock \emph{IEEE Transactions on Geoscience and Remote Sensing}, 64:\penalty0 1--20, 2026.

\bibitem[Tan and Le(2019)]{tan2019efficientnet}
Mingxing Tan and Quoc Le.
\newblock Efficientnet: Rethinking model scaling for convolutional neural networks.
\newblock In \emph{International conference on machine learning}, pages 6105--6114. PMLR, 2019.

\bibitem[{Taylor Geospatial Engine}(2025)]{fiboa2025}
{Taylor Geospatial Engine}.
\newblock Field boundaries for agriculture (fiboa) — specification, tools and open data, 2025.

\bibitem[Tolan et~al.(2024)Tolan, Yang, Nosarzewski, Couairon, Vo, Brandt, Spore, Majumdar, Haziza, Vamaraju, et~al.]{tolan2024very}
Jamie Tolan, Hung-I Yang, Benjamin Nosarzewski, Guillaume Couairon, Huy~V Vo, John Brandt, Justine Spore, Sayantan Majumdar, Daniel Haziza, Janaki Vamaraju, et~al.
\newblock {Very high resolution canopy height maps from RGB imagery using self-supervised vision transformer and convolutional decoder trained on aerial lidar}.
\newblock \emph{Remote Sensing of Environment}, 300:\penalty0 113888, 2024.

\bibitem[Tseng et~al.(2025)Tseng, Fuller, Reil, Herzog, Beukema, Bastani, Green, Shelhamer, Kerner, and Rolnick]{tseng2025galileolearningglobal}
Gabriel Tseng, Anthony Fuller, Marlena Reil, Henry Herzog, Patrick Beukema, Favyen Bastani, James~R Green, Evan Shelhamer, Hannah Kerner, and David Rolnick.
\newblock Galileo: Learning global \& local features of many remote sensing modalities.
\newblock In \emph{Proceedings of the 42nd International Conference on Machine Learning}, pages 60280--60300. PMLR, 2025.

\bibitem[Waldner and Diakogiannis(2020)]{WALDNER2020111741}
François Waldner and Foivos~I. Diakogiannis.
\newblock Deep learning on edge: Extracting field boundaries from satellite images with a convolutional neural network.
\newblock \emph{Remote Sensing of Environment}, 245:\penalty0 111741, 2020.

\bibitem[Waldner et~al.(2021)Waldner, Diakogiannis, Batchelor, Ciccotosto-Camp, Cooper-Williams, Herrmann, Mata, and Toovey]{rs13112197}
Fran{\c{c}}ois Waldner, Foivos~I. Diakogiannis, Kathryn Batchelor, Michael Ciccotosto-Camp, Elizabeth Cooper-Williams, Chris Herrmann, Gonzalo Mata, and Andrew Toovey.
\newblock {Detect, Consolidate, Delineate: Scalable Mapping of Field Boundaries Using Satellite Images}.
\newblock \emph{Remote Sensing}, 13\penalty0 (11), 2021.

\bibitem[Wang et~al.(2022)Wang, Waldner, and Lobell]{rs14225738}
Sherrie Wang, Fran{\c{c}}ois Waldner, and David~B. Lobell.
\newblock {Unlocking Large-Scale Crop Field Delineation in Smallholder Farming Systems with Transfer Learning and Weak Supervision}.
\newblock \emph{Remote Sensing}, 14\penalty0 (22), 2022.

\bibitem[Wang et~al.(2023)Wang, Shu, Han, Yang, Gordon, Wang, and Xu]{electronics12051156}
Xuying Wang, Lei Shu, Ru Han, Fan Yang, Timothy Gordon, Xiaochan Wang, and Hongyu Xu.
\newblock A survey of farmland boundary extraction technology based on remote sensing images.
\newblock \emph{Electronics}, 12\penalty0 (5), 2023.

\bibitem[Wang et~al.(2024{\natexlab{a}})Wang, Albrecht, Braham, Liu, Xiong, and Zhu]{wang2023decur}
Yi Wang, Conrad~M. Albrecht, Nassim Ait~Ali Braham, Chenying Liu, Zhitong Xiong, and Xiao~Xiang Zhu.
\newblock Decoupling common and unique representations for multimodal self-supervised learning.
\newblock In \emph{Eur. Conf. Comput. Vis.}, pages 286--303, 2024{\natexlab{a}}.

\bibitem[Wang et~al.(2024{\natexlab{b}})Wang, Albrecht, and Zhu]{wang2024multilabelguidedsoftcontrastive}
Yi Wang, Conrad~M Albrecht, and Xiao~Xiang Zhu.
\newblock {Multi-Label Guided Soft Contrastive Learning for Efficient Earth Observation Pretraining}.
\newblock In \emph{IGARSS 2024 - 2024 IEEE International Geoscience and Remote Sensing Symposium}, pages 7568--7571, 2024{\natexlab{b}}.

\bibitem[Watkins and {van Niekerk}(2019)]{WATKINS2019294}
Barry Watkins and Adriaan {van Niekerk}.
\newblock {A Comparison of Object-Based Image Analysis Approaches for Field Boundary Delineation Using Multi-Temporal Sentinel-2 Imagery}.
\newblock \emph{Computers and Electronics in Agriculture}, 158:\penalty0 294--302, 2019.

\bibitem[Wellington and Renzullo(2021)]{rs13071300}
Michael~J. Wellington and Luigi~J. Renzullo.
\newblock {High-Dimensional Satellite Image Compositing and Statistics for Enhanced Irrigated Crop Mapping}.
\newblock \emph{Remote Sensing}, 13\penalty0 (7), 2021.

\bibitem[Xiao et~al.(2018)Xiao, Liu, Zhou, Jiang, and Sun]{xiao2018unifiedperceptualparsingscene}
Tete Xiao, Yingcheng Liu, Bolei Zhou, Yuning Jiang, and Jian Sun.
\newblock {Unified Perceptual Parsing for Scene Understanding}.
\newblock In \emph{European Conference on Computer Vision}. Springer, 2018.

\bibitem[Xie et~al.(2021)Xie, Wang, Yu, Anandkumar, Alvarez, and Luo]{xie2021segformer}
Enze Xie, Wenhai Wang, Zhiding Yu, Anima Anandkumar, Jose~M Alvarez, and Ping Luo.
\newblock Segformer: Simple and efficient design for semantic segmentation with transformers.
\newblock \emph{Advances in neural information processing systems}, 34:\penalty0 12077--12090, 2021.

\bibitem[Xiong et~al.(2024)Xiong, Wang, Zhang, Stewart, Hanna, Borth, Papoutsis, Saux, Camps-Valls, and Zhu]{xiong2024neural}
Zhitong Xiong, Yi Wang, Fahong Zhang, Adam~J Stewart, Jo{\"e}lle Hanna, Damian Borth, Ioannis Papoutsis, Bertrand~Le Saux, Gustau Camps-Valls, and Xiao~Xiang Zhu.
\newblock {Neural Plasticity-Inspired Foundation Model for Observing the Earth Crossing Modalities}.
\newblock \emph{arXiv preprint arXiv:2403.15356}, 2024.

\bibitem[Yu et~al.(2021)Yu, Shi, Wei, Ren, Ye, and Tan]{DBLP:journals/corr/abs-2112-11037}
Xiaodong Yu, Dahu Shi, Xing Wei, Ye Ren, Tingqun Ye, and Wenming Tan.
\newblock {SOIT: Segmenting Objects with Instance-Aware Transformers}.
\newblock \emph{CoRR}, abs/2112.11037, 2021.

\bibitem[Zaheer et~al.(2017)Zaheer, Kottur, Ravanbakhsh, P{\'o}czos, Salakhutdinov, and Smola]{zaheer2017deepsets}
Manzil Zaheer, Satwik Kottur, Siamak Ravanbakhsh, Barnab{\'a}s P{\'o}czos, Ruslan Salakhutdinov, and Alexander~J. Smola.
\newblock Deep sets.
\newblock In \emph{Advances in Neural Information Processing Systems (NeurIPS)}, 2017.

\bibitem[Zhang(2019)]{zhang2019making}
Richard Zhang.
\newblock {Making Convolutional Networks Shift-Invariant Again}.
\newblock In \emph{Proceedings of the 36th International Conference on Machine Learning}, pages 7324--7334. PMLR, 2019.

\bibitem[Zou et~al.(2023)Zou, Xiao, Yu, Li, and Lee]{zou2023delving}
Xueyan Zou, Fanyi Xiao, Zhiding Yu, Yuheng Li, and Yong~Jae Lee.
\newblock {Delving Deeper into Anti-Aliasing in ConvNets}.
\newblock \emph{International Journal of Computer Vision}, 131\penalty0 (1):\penalty0 67--81, 2023.

\bibitem[Zvonkov et~al.(2023)Zvonkov, Tseng, Nakalembe, and Kerner]{Zvonkov_Tseng_Nakalembe_Kerner_2023}
Ivan Zvonkov, Gabriel Tseng, Catherine Nakalembe, and Hannah Kerner.
\newblock {OpenMapFlow: A Library for Rapid Map Creation with Machine Learning and Remote Sensing Data}.
\newblock \emph{Proceedings of the AAAI Conference on Artificial Intelligence}, 37\penalty0 (12):\penalty0 14655--14663, 2023.

\end{thebibliography}
}

\clearpage
\setcounter{page}{1}
\maketitlesupplementary

\appendix

\begin{figure*}[htpb!]
  \centering
  \includegraphics[width=\linewidth]{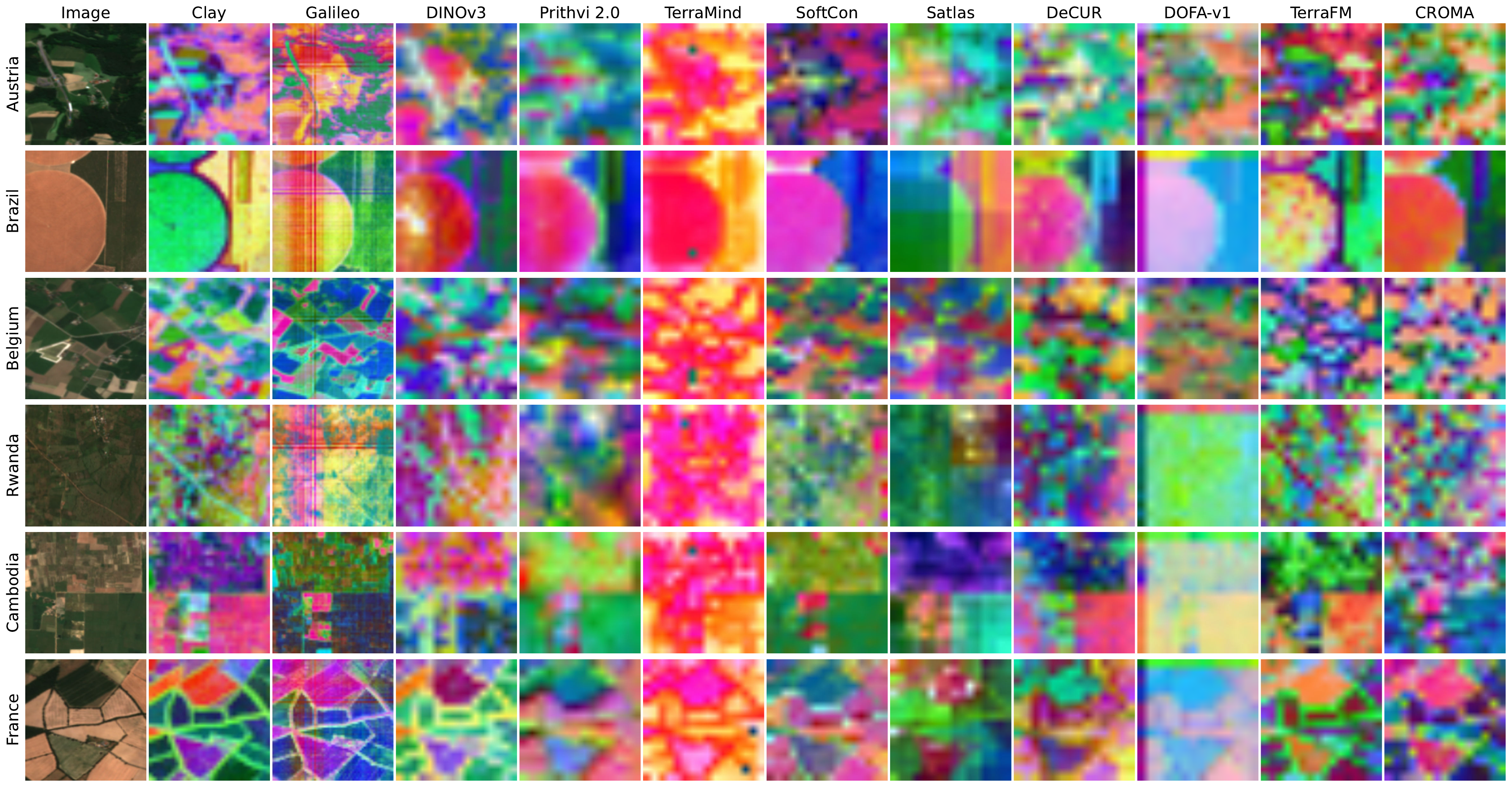}
  \caption{PCA visualization of frozen GFM encoder features for a representative subset of image examples. The top 3 principal components of patch embeddings are displayed as RGB channels. Clay ($8\times8$ patch sizes) and Galileo ($4\times4$ patch size) capture finer spatial structure and more distinct field boundaries compared to models using $16\times16$ patches, demonstrating how tokenization granularity affects feature quality for segmentation tasks. See Table~\ref{tab:1d_conv_gfm_results} for quantitative performance.}
  \label{fig:gfm_features}
\end{figure*}

\section{Extended Experimental Details}
\label{supp:extended_experiments}
\paragraph{Details on instance and panoptic segmentation model baselines.} Delineate Anything \cite{lavreniuk2025delineateanything} is based on Ultralytics' YOLOv11-seg and fine-tuned on FBIS-22M, a dataset of RGB images from multiple remote sensing sources (Sentinel-2, Planet, Maxar, Pleiades, orthophotos) over 9 European countries with spatial resolution 0.25-10m. For Delineate Anything, we perform a 1st-99th percentile normalization following the pretraining dataset norms. SAM \cite{kirillov2023sam} is a promptable instance segmentation model pretrained on natural images, which we assessed in both zero-shot and fine-tuned settings as described in the methods section. Mask2Former (M2F) \cite{cheng2022masked} is a universal segmentation architecture capable of semantic, instance, and panoptic segmentation depending on training configuration; we adapted M2F with a Swin-S backbone to handle the 8-channel RGBN bitemporal input and train it on the panoptic task, which jointly predicts individual field instances (things) and background land cover classes (stuff).
Note that SAM and Mask2Former were trained without presence-only label masking--a data preprocessing strategy used by all semantic baselines that filters out ambiguous background regions in partially-labeled countries. The inability to implement this masking for instance models (due to fundamental differences in how instance segmentation models handle training objectives) means these models faced a harder optimization landscape, being penalized for predicting fields in regions that may contain unlabeled fields.

\paragraph{RGB-only comparison with Delineate Anything.}
To provide a fairer comparison with Delineate Anything (DelAny)~\cite{lavreniuk2025delineateanything}, we trained new PRUE models with EfficientNet-B3 (EF3) backbones on RGB-only data from a single time step. This resulted in object F1 scores of $0.38\pm0.06$ (window A) and $0.37\pm0.06$ (window B), both higher than DelAny's performance despite using the same RGB-only input. This demonstrates that the performance gap between PRUE and DelAny is not attributable to FTW's additional spectral channel (NIR), but rather reflects differences in model design and training data diversity.

\paragraph{SAM2 evaluation.}
We evaluated SAM2 in a zero-shot setting on the window A RGB bands, resulting in a pixel IoU of $0.31$ and an object F1 of $0.07$. SAM2 is designed for video segmentation where it expects continuous video frames in which objects move or deform slightly with strong appearance consistency~\cite{ravi2024sam2segmentimages}. This differs fundamentally from the FTW setting, which provides two snapshots separated by months that capture significant phenological changes. This creates multiple failure modes: (1) appearance shifts between seasons violate SAM2's visual consistency assumption, and (2) fields do not ``move'' between timestamps---they transform---leaving SAM2's optical flow and correspondence mechanisms without useful signal. 

\paragraph{Geospatial foundation models (GFMs).}The FTW benchmark provides four bands (RGB and NIR)~\cite{kerner2025fields}, which is fewer than the multi-spectral inputs used by most GFMs. Since many GFMs expect 8 to 13 Sentinel-2 bands, we used GFMs evaluation wrapper published in Galileo codebase~\cite{tseng2025galileolearningglobal} to correctly prepare inputs for Galileo \cite{tseng2025galileolearningglobal}, CROMA \cite{fuller2024croma}, SoftCon \cite{wang2024multilabelguidedsoftcontrastive}, Prithvi 2.0 \cite{szwarcman2025prithvieo20versatilemultitemporalfoundation}, DOFA-v1 \cite{xiong2024neural}, DeCUR \cite{wang2023decur}, and Satlas \cite{bastani2023satlaspretrainlargescaledatasetremote}. This wrapper allowed us to (1) construct the band set expected by each model (applying mask where applicable), (2) impute missing channels in a model-consistent manner, and (3) apply each model's required normalization or standardization using its original training statistics.
For TerraFM~\cite{danish2025terrafmscalablefoundationmodel}, we assign zeros to all missing spectral bands. DINOv3~\cite{simeoni2025dinov3} operates exclusively on RGB inputs, so the FTW RGB bands are passed directly without modification. For Clay~\cite{clay2025model} and TerraMind~\cite{jakubik2025terramindlargescalegenerativemultimodality}, which are designed to handle partially missing spectral information, we provide the available four-band input with the appropriate normalization for each model. Patch-level embeddings are extracted from each pretrained GFM independently for the two temporal windows defined in FTW.

\paragraph{GFM feature fusion and decoding.} 
The patch embeddings from the two temporal windows are fused by first concatenating them along the feature dimension and passing the result through a three-layer MLP. Our objective is to evaluate the representational quality of the frozen GFM features themselves. A common evaluation strategy adopted for this objective is linear probing. However, we argue that a single linear transformation is often too limited to fully assess the spatial and contextual information encoded in the pretrained features. Conversely, using a specialized model such as U\text{-}Net~\cite{ronneberger2015unetconvolutionalnetworksbiomedical} at this stage would primarily test the ability of the decoder rather than the underlying GFM embeddings. To strike a balance between these two extremes, we employ a simple decoder that provides moderate flexibility through a $3{\times}3$ projection layer, two residual refinement blocks, and a multi-scale convolutional module, followed by pixel-shuffle upsampling. Table~\ref{tab:baselines_results} reports the results obtained with our convolutional decoder, and Table~\ref{tab:1d_conv_gfm_results} provides the complementary $1{\times}1$ convolution linear-probing results.

\begin{table*}[t]
\centering
\caption{GFM linear probing results using a lightweight decoder (1$\times$1 convolution + bilinear upsampling), sorted by object-level F1. Clay (ViT-Large, 8$\times$8 patches) and Galileo (ViT-Base, 4$\times$4 patches) outperform other GFMs that use coarser 16$\times$16 patch sizes, due to the finer patch resolutions as well as techniques intentionally designed to handle missing spectral bands.}
\label{tab:1d_conv_gfm_results}
\resizebox{0.7\textwidth}{!}{%
\begin{tabular}{@{}llccc@{\hspace{4pt}}ccccc@{}}
\toprule
\multirow{2}{*}{\textbf{Model}} & \multirow{2}{*}{\textbf{Backbone}} &
\multicolumn{3}{c}{\textbf{Pixel-level}} &
\multicolumn{5}{c}{\textbf{Object-level}} \\
\cmidrule(lr){3-5} \cmidrule(lr){6-10}
& & IoU $\uparrow$ & Prec $\uparrow$ & Recall $\uparrow$
& Prec $\uparrow$ & Recall $\uparrow$ & F1 $\uparrow$
& AP$_{0.5:0.95}$ $\uparrow$ & AP$_{0.5}$ $\uparrow$ \\
\midrule

Clay      & ViT-Large & \maxval{0.56} & \secondval{0.88} & \maxval{0.60} & \secondval{0.22} & \secondval{0.16} & \maxval{0.18} & \secondval{0.07} & \secondval{0.17} \\
Galileo   & ViT-Base  & \secondval{0.53} & 0.83 & \secondval{0.59} & 0.11 & \maxval{0.19} & \secondval{0.13} & \maxval{0.08} & \maxval{0.18} \\
DINOv3    & ViT-Large & 0.47 & \maxval{0.89} & 0.50 & \maxval{0.25} & 0.09 & 0.12 & 0.03 & 0.08 \\
Prithvi 2.0 & ViT-Large & 0.44 & 0.84 & 0.48 & 0.20 & 0.06 & 0.10 & 0.02 & 0.06 \\
TerraMind & ViT-Base & 0.44 & 0.85 & 0.47 & 0.19 & 0.07 & 0.10 & 0.02 & 0.06 \\
SoftCon   & ViT-Small & 0.41 & 0.83 & 0.46 & 0.16 & 0.05 & 0.07 & 0.01 & 0.04 \\
Satlas    & Swin-Tiny & 0.39 & 0.74 & 0.45 & 0.13 & 0.04 & 0.07 & 0.01 & 0.03 \\
DeCUR     & ViT-Small & 0.42 & 0.80 & 0.46 & 0.15 & 0.04 & 0.07 & 0.01 & 0.03 \\
DOFA-v1   & ViT-Large & 0.39 & 0.77 & 0.44 & 0.14 & 0.04 & 0.06 & 0.01 & 0.03 \\
TerraFM   & ViT-Base & 0.44 & 0.85 & 0.48 & 0.17 & 0.06 & 0.09 & 0.02 & 0.05 \\
CROMA     & ViT-Base & 0.42 & 0.85 & 0.46 & 0.18 & 0.05 & 0.08 & 0.02 & 0.05 \\

\bottomrule
\end{tabular}
}
\end{table*}

\section{Extended Results}
\label{supp:extended_ablation}

\begin{table*}
\centering
\caption{Ablation results for controlled experiments on the FTW test set (excluding presence-only countries) in which each row varies a single design choice (data augmentations, class weighting, encoder, loss function, or architecture). \textbf{Bold} indicates best performance, \underline{underline} indicates second-best.}
\resizebox{\textwidth}{!}{%
\begin{tabular}{@{}llccccccccc@{}}
\toprule
   \multirow{2}{*}{\textbf{Category}} &
   \multirow{2}{*}{\textbf{Ablation}} &
  \multicolumn{2}{c}{\textbf{Performance}} &
  \multicolumn{2}{c}{\textbf{Input order}} &
  \multicolumn{2}{c}{\textbf{Brightness}} &
  \multicolumn{2}{c}{\textbf{Scale}} &
  \multicolumn{1}{c}{\textbf{Agree.}} \\

  \cmidrule(lr){3-4}
  \cmidrule(lr){5-6}
  \cmidrule(lr){7-8}
  \cmidrule(lr){9-10}
  \cmidrule(l){11-11}

 & &
  Object F1 $\uparrow$ &
  Pixel IoU $\uparrow$ &
  F1$|\Delta|\downarrow$ &
  IoU$|\Delta|\downarrow$ &
  F1$|\Delta|\downarrow$ &
  IoU$|\Delta|\downarrow$ &
  F1$|\Delta|\downarrow$ &
  IoU$|\Delta|\downarrow$ &
  Avg$\uparrow$ \\
\midrule

 & FTW-v1 &
  0.39 $\pm$ 0.08 &
  0.68 $\pm$ 0.08 &
  0.07 &
  0.11 &
  0.04 &
  0.05 &
  0.17 &
  0.08 &
  0.93 \\
\midrule

Data augs & Brightness &
  0.39 $\pm$ 0.08 &
  0.68 $\pm$ 0.08 &
  0.07 &
  0.09 &
  \textbf{0.00} &
  \textbf{0.00} &
  0.14 &
  0.05 &
  0.93 \\

Data augs & Resize &
  0.38 $\pm$ 0.08 &
  0.67 $\pm$ 0.09 &
  0.07 &
  0.10 &
  0.04 &
  0.05 &
  \textbf{0.00} &
  \underline{0.01} &
  0.94 \\

Data augs & Brightness+Resize &
  0.38 $\pm$ 0.08 &
  0.66 $\pm$ 0.09 &
  0.06 &
  0.10 &
  0.03 &
  \underline{0.03} &
  \textbf{0.00} &
  0.02 &
  0.95 \\

Data augs & Channel shuffle &
  0.39 $\pm$ 0.07 &
  0.68 $\pm$ 0.09 &
  \textbf{0.00} &
  \textbf{0.00} &
  0.05 &
  0.05 &
  0.18 &
  0.09 &
  0.94 \\

\midrule

Class weights & $\omega = 0.60$ &
  0.32 $\pm$ 0.06 &
  \underline{0.76 $\pm$ 0.06} &
  0.07 &
  0.13 &
  0.07 &
  0.07 &
  0.28 &
  0.19 &
  \underline{0.96} \\

Class weights & $\omega = 0.65$ &
  0.36 $\pm$ 0.06 &
  \underline{0.76 $\pm$ 0.06} &
  0.07 &
  0.11 &
  0.07 &
  0.07 &
  0.30 &
  0.17 &
  \underline{0.96} \\

Class weights & $\omega = 0.70$ &
  0.40 $\pm$ 0.06 &
  0.75 $\pm$ 0.06 &
  0.08 &
  0.12 &
  0.08 &
  0.07 &
  0.30 &
  0.14 &
  0.95 \\

Class weights & $\omega = 0.75$ &
  0.42 $\pm$ 0.06 &
  0.74 $\pm$ 0.07 &
  0.08 &
  0.11 &
  0.07 &
  0.07 &
  0.29 &
  0.13 &
  0.95 \\

Class weights & $\omega = 0.80$ &
  0.42 $\pm$ 0.07 &
  0.73 $\pm$ 0.06 &
  0.09 &
  0.12 &
  0.07 &
  0.07 &
  0.23 &
  0.11 &
  0.95 \\

Class weights & $\omega = 0.85$ &
  0.41 $\pm$ 0.07 &
  0.70 $\pm$ 0.08 &
  0.08 &
  0.10 &
  0.05 &
  0.06 &
  0.17 &
  0.08 &
  \underline{0.96} \\
\midrule

Encoder & EfficientNet-B4 &
  0.40 $\pm$ 0.07 &
  0.69 $\pm$ 0.09 &
  0.07 &
  0.09 &
  0.04 &
  0.05 &
  0.15 &
  0.06 &
  0.93 \\

Encoder & EfficientNet-B5 &
  0.41 $\pm$ 0.07 &
  0.70 $\pm$ 0.08 &
  0.07 &
  0.11 &
  0.04 &
  0.05 &
  0.16 &
  0.10 &
  0.93 \\

Encoder & EfficientNet-B6 &
  0.41 $\pm$ 0.07 &
  0.70 $\pm$ 0.08 &
  0.07 &
  0.11 &
  0.04 &
  0.05 &
  0.18 &
  0.14 &
  0.94 \\

Encoder & EfficientNet-B7 &
  0.42 $\pm$ 0.07 &
  0.71 $\pm$ 0.08 &
  0.07 &
  0.09 &
  0.04 &
  0.04 &
  0.21 &
  0.09 &
  0.94 \\

Encoder & MiT-B2 &
  0.39 $\pm$ 0.08 &
  0.67 $\pm$ 0.09 &
  0.08 &
  0.10 &
  \underline{0.02} &
  \underline{0.03} &
  0.13 &
  0.05 &
  0.95 \\

Encoder & MiT-B3 &
  0.39 $\pm$ 0.08 &
  0.67 $\pm$ 0.09 &
  0.08 &
  0.10 &
  \underline{0.02} &
  \underline{0.03} &
  0.13 &
  0.05 &
  0.95 \\

Encoder & MiT-B4 &
  0.39 $\pm$ 0.08 &
  0.68 $\pm$ 0.09 &
  0.07 &
  0.09 &
  \underline{0.02} &
  \underline{0.03} &
  0.16 &
  0.06 &
  0.94 \\

Encoder & MiT-B5 &
  0.38 $\pm$ 0.08 &
  0.67 $\pm$ 0.09 &
  0.08 &
  0.10 &
  \underline{0.02} &
  \underline{0.03} &
  0.12 &
  0.02 &
  0.95 \\

Encoder & ResNet-18 &
  0.35 $\pm$ 0.07 &
  0.67 $\pm$ 0.09 &
  0.08 &
  0.11 &
  0.04 &
  0.05 &
  0.14 &
  0.05 &
  0.93 \\

Encoder & VGG13-BN &
  0.38 $\pm$ 0.07 &
  0.69 $\pm$ 0.08 &
  0.08 &
  0.10 &
  0.04 &
  0.05 &
  0.27 &
  0.20 &
  0.94 \\
\midrule

Learning rate & 0.0001 &
  0.34 $\pm$ 0.07 &
  0.65 $\pm$ 0.09 &
  0.05 &
  \underline{0.07} &
  0.03 &
  0.04 &
  0.15 &
  0.08 &
  0.89 \\

Learning rate & 0.0003 &
  0.37 $\pm$ 0.07 &
  0.66 $\pm$ 0.09 &
  0.06 &
  0.08 &
  0.04 &
  0.04 &
  0.17 &
  0.10 &
  0.90 \\

Learning rate & 0.003 &
  0.39 $\pm$ 0.08 &
  0.68 $\pm$ 0.09 &
  0.08 &
  0.10 &
  0.06 &
  0.08 &
  0.17 &
  0.08 &
  0.95 \\

Learning rate & 0.01 &
  0.39 $\pm$ 0.08 &
  0.68 $\pm$ 0.09 &
  0.08 &
  0.11 &
  0.06 &
  0.06 &
  0.18 &
  0.08 &
  0.95 \\

Learning rate & 0.03 &
  0.37 $\pm$ 0.08 &
  0.67 $\pm$ 0.09 &
  0.09 &
  0.10 &
  0.10 &
  0.13 &
  0.16 &
  0.07 &
  0.94 \\
\midrule

Loss function & CE (w/ Edge Agreement) &
  0.39 $\pm$ 0.08 &
  0.68 $\pm$ 0.09 &
  0.07 &
  0.10 &
  0.04 &
  0.05 &
  0.15 &
  0.07 &
  0.94 \\

Loss function & CE + Dice &
  0.41 $\pm$ 0.07 &
  0.70 $\pm$ 0.08 &
  0.07 &
  0.10 &
  0.04 &
  0.05 &
  0.20 &
  0.08 &
  0.93 \\

Loss function & CE + Dice (no class weights) &
  0.38 $\pm$ 0.07 &
  \textbf{0.77 $\pm$ 0.06} &
  0.08 &
  0.11 &
  0.06 &
  0.05 &
  0.29 &
  0.15 &
  0.95 \\

Loss function & CE + FTNMT &
  0.41 $\pm$ 0.07 &
  0.70 $\pm$ 0.08 &
  0.08 &
  0.11 &
  0.04 &
  0.05 &
  0.21 &
  0.08 &
  0.94 \\

Loss function & CE (no class weights) &
  0.24 $\pm$ 0.06 &
  \textbf{0.77 $\pm$ 0.06} &
  0.05 &
  0.11 &
  0.05 &
  0.06 &
  0.15 &
  0.14 &
  \underline{0.96} \\

Loss function & Dice &
  0.42 $\pm$ 0.07 &
  \underline{0.76 $\pm$ 0.07} &
  0.08 &
  0.13 &
  0.06 &
  0.06 &
  0.31 &
  0.16 &
  \underline{0.96} \\

Loss function & Dice (w/ Edge Agreement) &
  0.42 $\pm$ 0.07 &
  \textbf{0.77 $\pm$ 0.06} &
  0.08 &
  0.13 &
  0.07 &
  0.07 &
  0.32 &
  0.17 &
  0.95 \\

Loss function & Focal &
  0.18 $\pm$ 0.05 &
  0.75 $\pm$ 0.06 &
  0.04 &
  0.10 &
  0.04 &
  0.06 &
  0.15 &
  0.21 &
  \underline{0.96} \\

Loss function & FTNMT &
  0.38 $\pm$ 0.06 &
  0.79 $\pm$ 0.06 &
  0.10 &
  0.14 &
  0.06 &
  0.07 &
  0.29 &
  0.17 &
  0.93 \\

Loss function & Local Tversky &
  0.45 $\pm$ 0.07 &
  0.74 $\pm$ 0.07 &
  0.10 &
  0.13 &
  0.05 &
  0.05 &
  0.31 &
  0.16 &
  0.94 \\

Loss function & LogCosh Dice &
  0.44 $\pm$ 0.07 &
  \textbf{0.77 $\pm$ 0.06} &
  0.09 &
  0.13 &
  0.06 &
  0.06 &
  0.33 &
  0.20 &
  0.94 \\

Loss function & LogCosh Dice + CE &
  0.39 $\pm$ 0.08 &
  0.68 $\pm$ 0.08 &
  0.08 &
  0.10 &
  0.04 &
  0.05 &
  0.14 &
  0.06 &
  0.93 \\

Loss function & Tversky &
  0.43 $\pm$ 0.07 &
  \underline{0.76 $\pm$ 0.06} &
  0.09 &
  0.12 &
  0.06 &
  0.05 &
  0.34 &
  0.15 &
  0.95 \\

Loss function & Tversky + CE &
  0.41 $\pm$ 0.07 &
  0.71 $\pm$ 0.08 &
  0.07 &
  0.10 &
  0.05 &
  0.06 &
  0.20 &
  0.10 &
  0.92 \\

\midrule

Architecture & FCN &
  0.14 $\pm$ 0.03 &
  0.60 $\pm$ 0.08 &
  \underline{0.04} &
  0.09 &
  0.03 &
  0.07 &
  0.10 &
  \textbf{0.00} &
  \textbf{0.99} \\

Architecture & FCSiam &
  0.40 $\pm$ 0.07 &
  0.69 $\pm$ 0.08 &
  \textbf{0.00} &
  \textbf{0.00} &
  0.05 &
  0.06 &
  0.23 &
  0.10 &
  0.92 \\

Architecture & UNETR &
  0.37 $\pm$ 0.07 &
  0.69 $\pm$ 0.08 &
  0.08 &
  0.10 &
  0.04 &
  0.04 &
  0.27 &
  0.18 &
  0.94 \\

Architecture & UPerNet &
  0.34 $\pm$ 0.08 &
  0.64 $\pm$ 0.10 &
  0.07 &
  0.09 &
  0.03 &
  0.04 &
  0.13 &
  0.04 &
  0.91 \\
\midrule

 Combination & FCSiam Combo & %
  0.44 $\pm$ 0.07 &
  0.75 $\pm$ 0.07 &
  \textbf{0.00} &
  \textbf{0.00} &
  0.04 &
  0.04 &
  0.05 &
  0.02 &
  0.94 \\

  Combination & U-Net LogCosh Dice Augs EN-B3 &
  0.42 $\pm$ 0.07 &
  0.74 $\pm$ 0.07 &
  \textbf{0.00} &
  \textbf{0.00} &
  \textbf{0.00} &
  \textbf{0.00} &
  \underline{0.01} &
  \underline{0.01} &
  0.94 \\

  Combination & U-Net LogCosh Dice 0.75-weight Augs EN-B3 &
  0.43 $\pm$ 0.07 &
  0.74 $\pm$ 0.07 &
  \textbf{0.00} &
  \textbf{0.00} &
  \textbf{0.00} &
  \textbf{0.00} &
  0.02 &
  \textbf{0.00} &
  0.94 \\

Combination & U-Net LogCosh Dice 0.75-weight Augs EN-B5 &
  \underline{0.46 $\pm$ 0.07} &
  0.75 $\pm$ 0.07 &
  \textbf{0.00 }&
  \textbf{0.00} &
  \textbf{0.00} &
  \textbf{0.00} &
  \underline{0.01} &
  \underline{0.01} &
  0.94 \\

 Combination & \textbf{PRUE} &
  \textbf{0.47 $\pm$ 0.07} &
  \underline{0.76 $\pm$ 0.08} &
  \textbf{0.00} &
  \textbf{0.00} &
  \textbf{0.00} &
  \textbf{0.00} &
  \underline{0.01} &
  \underline{0.01} &
  0.95 \\

\bottomrule 
\end{tabular}%
}

\label{tab:ablation_supp}
\end{table*}

\paragraph{Our design choice is a result of extensive ablations.} 
Table~\ref{tab:ablation_supp} shows that boundary weighting, loss function, and targeted augmentations have the strongest impact on performance. Moderate class weights ($\omega \approx 0.75$) and losses such as LogCosh Dice, Tversky, and Local Tversky consistently improve object F1 and pixel IoU, while brightness, scale, and channel-shuffle augmentations provide additional robustness. Larger EfficientNet backbones slightly enhance results, and combining these components in PRUE yields the highest accuracy and boundary agreement across all metrics.

\begin{figure}[h]
\centering
\includegraphics[width=0.99\linewidth]{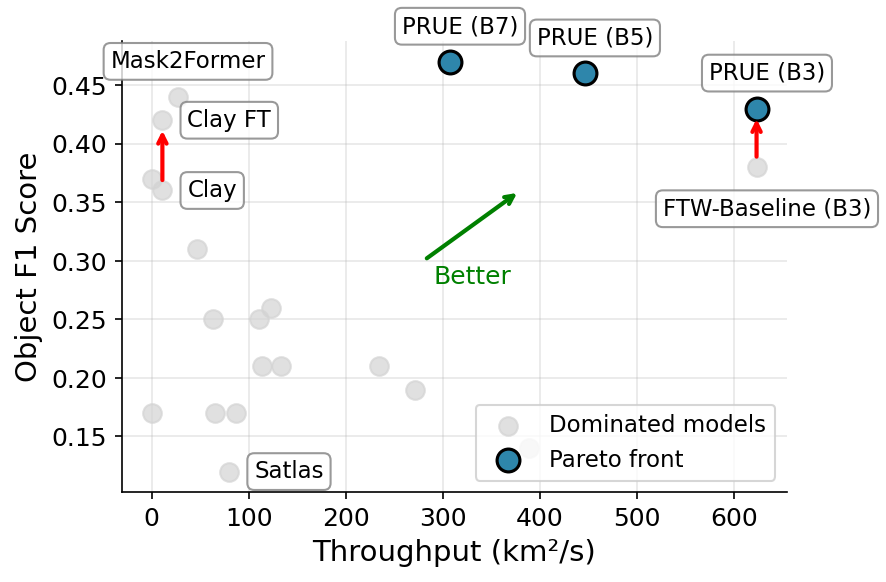}
\caption{Pareto front between Object F1 and Throughput for all Table~1 models, including PRUE-EF (B3/B5) throughput. PRUE-EF3 is a Pareto improvement over the FTW baseline. The PRUE family forms a substantially stronger Pareto frontier than any prior model.}
\label{fig:pareto}
\end{figure}

\paragraph{Accuracy--throughput trade-off across model configurations.}
To explicitly summarize the accuracy--cost trade-off over all models, Figure~\ref{fig:pareto} shows the Pareto front between object F1 and throughput. Supplemental Table~\ref{tab:ablation_supp} compares our methodology against the baseline using the same backbone (EfficientNet-B3): the ``U-Net LogCosh Dice 0.75-weight Augs EN-B3'' row shows that PRUE with an EF3 backbone achieves an object F1 of $0.43\pm0.07$ and field IoU of $0.74\pm0.07$, compared to PRUE-EF7 which achieves $0.47\pm0.07$ and $0.76\pm0.08$, respectively. PRUE-EF3 is a Pareto improvement over the FTW baseline, meaning it achieves higher accuracy \textit{without} sacrificing throughput. The PRUE family forms a substantially stronger Pareto frontier than any prior model and shows a clear throughput vs.\ performance trade-off. As a practical reference, an EF3 backbone can process the entire land area of the Earth in approximately 66 hours on a single V100 GPU, while EF7 would require approximately 134 hours. 

\paragraph{Full fine-tuning of Clay.}
To assess whether the performance gap between GFMs and PRUE could be closed with end-to-end training, we fully fine-tuned the best-performing GFM, Clay, selecting the learning rate from $\{1,3\} \times 10^{\{-5,-4,-3\}}$ based on the best object F1 on the validation set. Full fine-tuning increased Clay's object F1 from $0.36$ to $0.42$ (see Figure~\ref{fig:pareto}) and pixel IoU from $0.67$ to $0.73$. While these represent meaningful improvements over frozen-encoder decoding, both metrics remain below PRUE ($0.47$ object F1, $0.76$ pixel IoU), and Clay's throughput is substantially lower (Clay: 11~km$^2$/s vs.\ PRUE: 307~km$^2$/s). This confirms that the performance gap is not solely attributable to the frozen-encoder evaluation protocol, but reflects fundamental differences in spatial resolution and architectural suitability for field boundary delineation. 

\paragraph{1D convolution did not capture field boundary complexity.} Across all GFM experiments, we observed that decoding using a single $1\times1$ convolution followed by bilinear upsampling produced coarse, low-fidelity boundaries compared to our convolutional decoder head. Table~\ref{tab:1d_conv_gfm_results} shows these linear probing results for all GFMs.
Among the GFMs, Clay and Galileo exhibit notably stronger performance. Both models use smaller token patch sizes ($8{\times}8$ for Clay and $4{\times}4$ for Galileo), compared to most of the other evaluated models that operate at a patch size of $16{\times}16$. These finer patch resolutions produce higher-granularity spatial features that align more naturally with field-level segmentation. In addition, both models robustly handle missing spectral channels: Clay is explicitly trained with missing-band augmentation, and Galileo incorporates consistent band masking and normalization statistics for partially observed inputs.

\begin{figure*}[h!]
  \centering
  \includegraphics[width=\linewidth]{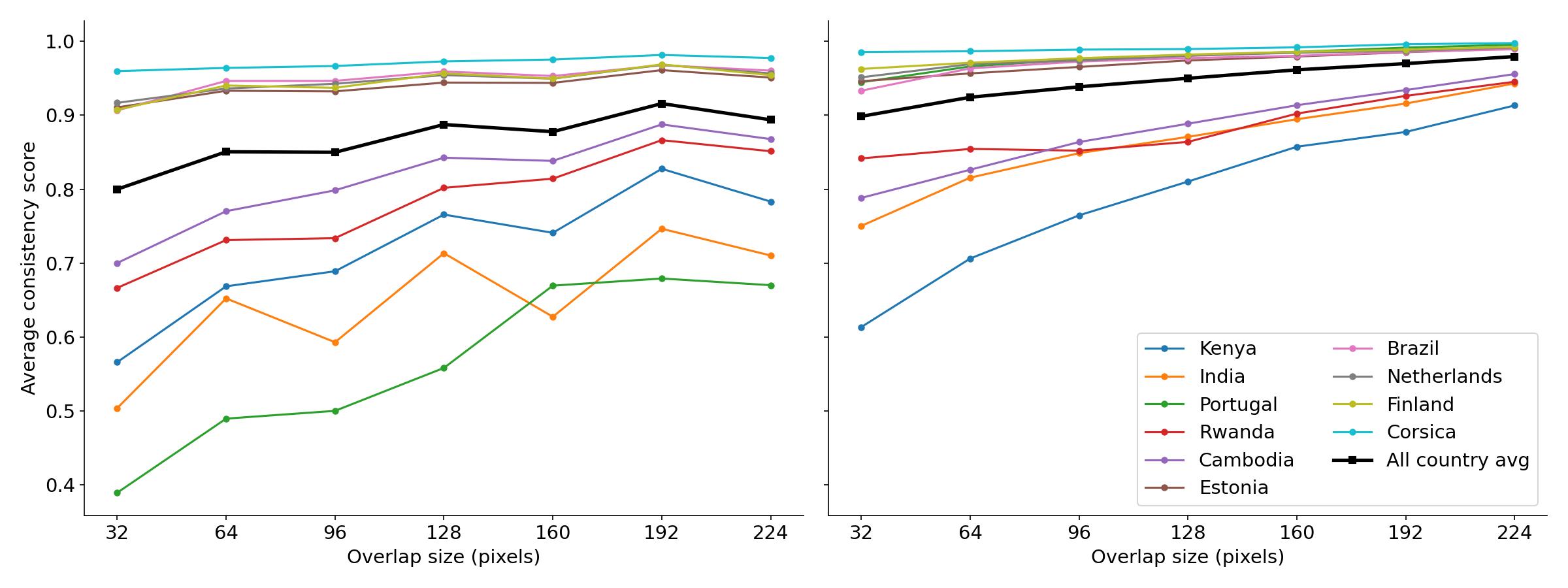}
  \caption{\textbf{Consistency dependence on overlap size.}
  Consistency scores as a function of overlap window size for the FTW baseline (left) and PRUE (right). Larger overlaps measure intrinsic translation sensitivity, while smaller overlaps additionally capture context dependency. Shown: top 5 and bottom 5 countries by mean consistency for each model. PRUE achieves higher consistency across all countries and overlap sizes, with particularly strong improvements in challenging regions. Hard countries (Kenya, India) show low consistency even at large overlaps, while well-represented regions (Finland, Netherlands) have high consistency, suggesting consistency metrics can identify out-of-distribution samples at inference time.
  }
  \label{fig:consistency_comp}
\end{figure*}

\paragraph{Consistency varies by geography and overlap window size.} Figure~\ref{fig:consistency_comp} illustrates how consistency changes with the size of the overlapping region between the four corner crops used to measure translation sensitivity. When the overlap window is large, approaching the patch size, consistency primarily reflects intrinsic translation sensitivity, which captures the model's architectural tendency to produce different outputs for identical content at different spatial positions. When the overlap window is small, consistency also reflects context dependency, indicating how much the model's predictions vary based on surrounding image content. 

PRUE achieved two to three times higher consistency than the FTW baseline across all overlap window sizes, demonstrating improvements in both intrinsic translation robustness, an architectural property, and reduced context dependency, a learned behavior. Consistency varied substantially across countries and correlates with test set difficulty. For example, our best model shows $>40$\% pixel disagreement in Kenya at 224-pixel context shifts, compared to $<20$\% in Switzerland. This suggests that consistency metrics may serve as an out-of-distribution detection signal, where low consistency during inference could indicate that the model is operating on data that differs from its training distribution, and may thus require human review or model retraining.

\begin{figure}[h]
\centering
\includegraphics[width=0.99\linewidth]{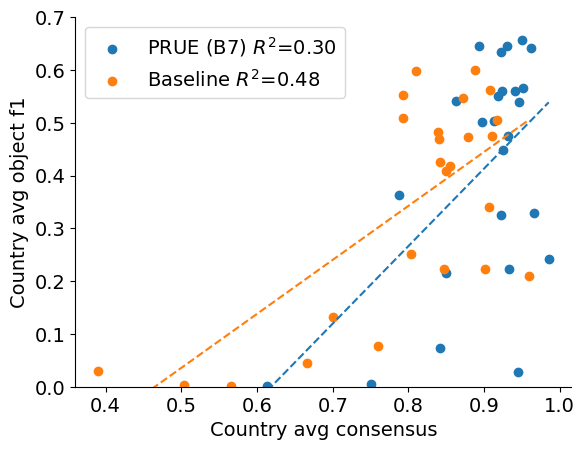}
\caption{\textbf{Country-level consistency vs. object F1.} Each point represents one FTW country. Dashed lines show linear fits for the FTW baseline (orange, R²=0.48) and PRUE-EF7 (blue, R²=0.30). Consistency is weakly correlated with performance, suggesting it may serve as a partial out-of-distribution signal but not a reliable sole predictor of model accuracy.}
\label{fig:consensus_vs_f1}
\end{figure}

\paragraph{Consistency as a reliability signal.}
To quantify the relationship between consistency and performance, we computed the average consistency over test samples per country and examined its correlation with object F1 (See Figure\ref{fig:consensus_vs_f1}). 
We observe that consistency is weakly correlated with performance: the FTW baseline yields $R^2{=}0.48$ and PRUE-EF7 yields $R^2{=}0.30$. This indicates that consistency metrics explain \textit{some} drops in out-of-distribution (OOD) performance but not all---a model can have high consistency but poor performance. 
Consistency metrics may serve as one signal among several for OOD detection, but should not be used as a sole indicator of model reliability. In deployment, low consistency scores are most useful for flagging regions that exhibit gridding artifacts and spatial prediction instability, warranting human review. 

\section{Per-Country Evaluation}
\label{supp:country_eval}
We report full per-country metrics for all FTW countries with complete labels, excluding presence-only regions. For each country, we provide pixel IoU and object-level F1.

To represent the major model families, we evaluated the strongest-performing model from each category, as identified in Table~\ref{tab:baselines_results} of the main paper: the FTW baseline for semantic segmentation~\cite{kerner2025fields}, Mask2Former (M2F) with a Swin Small backbone for panoptic segmentation~\cite{cheng2022masked}, Clay with our convolutional decoder for geospatial foundation models~\cite{clay2025model}, and PRUE. All metrics were computed on each country's official FTW test split, following the evaluation protocol described in~\S\ref{subsec:model_metrics}.

Across regions, the results show consistent patterns. As shown in Table~\ref{tab:country_level_supp},  PRUE consistently achieved the highest or second-highest Pixel IoU and Object F1 across nearly all tested FTW countries, highlighting its robust and reliable performance across diverse geographies and agricultural systems.

\begin{table*}[ht!]
\centering
\caption{Per-country performance comparison for the top-performing models of each architecture family: FTW baseline (semantic), Mask2Former with Swin-S (instance/panoptic), Clay (frozen GFM), Clay-FT (finetuned GFM), and PRUE (our optimized semantic model). Bold indicates best performance per country. PRUE achieves the highest or second-highest pixel IoU scores across nearly all countries. All models evaluated using the protocol in Section~\ref{subsec:model_metrics} on presence/absence labeled countries only.}
\resizebox{0.78\textwidth}{!}{%
\begin{tabular}{@{}lccccc@{\hskip 10pt}ccccc@{}}
\toprule
\multirow{2}{*}{\textbf{Country}} &
\multicolumn{5}{c}{\textbf{Pixel IoU}} &
\multicolumn{5}{c}{\textbf{Object F1}} \\
\cmidrule(lr){2-6} \cmidrule(lr){7-11}
& FTW & M2F & Clay & Clay-FT & PRUE
& FTW & M2F & Clay & Clay-FT & PRUE \\
\midrule

Austria       & 0.71 & 0.74 & 0.69 & 0.76 & \maxval{0.78} & 0.41 & 0.38 & 0.39 & 0.47 & \maxval{0.50} \\
Belgium       & 0.75 & 0.78 & 0.73 & 0.79 & \maxval{0.82} & 0.57 & 0.57 & 0.57 & 0.62 & \maxval{0.66} \\
Cambodia      & 0.40 & 0.19 & 0.27 & 0.40 & \maxval{0.66} & 0.19 & 0.10 & 0.09 & 0.19 & \maxval{0.36} \\
Corsica       & 0.45 & \maxval{0.52} & 0.47 & 0.51 & 0.51 & 0.18 & 0.24 & 0.18 & 0.22 & \maxval{0.24} \\
Croatia       & 0.67 & 0.70 & 0.64 & 0.71 & \maxval{0.77} & 0.28 & 0.34 & 0.25 & 0.33 & \maxval{0.45} \\
Denmark       & 0.83 & 0.57 & 0.83 & \maxval{0.86} & \maxval{0.86} & 0.52 & 0.24 & 0.51 & 0.58 & \maxval{0.65} \\
Estonia       & 0.80 & 0.82 & 0.80 & 0.83 & \maxval{0.84} & 0.44 & \maxval{0.54} & 0.43 & 0.48 & \maxval{0.54} \\
Finland       & 0.83 & 0.85 & 0.81 & 0.85 & \maxval{0.87} & 0.55 & 0.54 & 0.53 & 0.59 & \maxval{0.64} \\
France        & 0.79 & 0.80 & 0.78 & 0.81 & \maxval{0.83} & 0.55 & 0.55 & 0.54 & 0.58 & \maxval{0.63} \\
Germany       & 0.79 & 0.77 & 0.78 & \maxval{0.80} & 0.79 & 0.41 & 0.44 & 0.39 & 0.42 & \maxval{0.47} \\
Latvia        & 0.81 & 0.84 & 0.81 & 0.84 & \maxval{0.85} & 0.44 & 0.54 & 0.44 & 0.49 & \maxval{0.56} \\
Lithuania     & 0.74 & 0.78 & 0.74 & 0.78 & \maxval{0.79} & 0.39 & 0.48 & 0.38 & 0.45 & \maxval{0.50} \\
Luxembourg    & 0.79 & 0.80 & 0.76 & 0.82 & \maxval{0.85} & 0.49 & 0.37 & 0.46 & 0.53 & \maxval{0.56} \\
Netherlands   & 0.75 & 0.78 & 0.74 & \maxval{0.81} & \maxval{0.81} & 0.48 & 0.51 & 0.48 & 0.54 & \maxval{0.57} \\
Portugal      & 0.12 & 0.21 & 0.23 & \maxval{0.37} & 0.10 & 0.03 & \maxval{0.08} & 0.04 & 0.07 & 0.03 \\
Slovakia      & 0.92 & 0.91 & 0.92 & \maxval{0.94} & \maxval{0.94} & 0.53 & 0.61 & 0.53 & 0.58 & \maxval{0.65} \\
Slovenia      & 0.58 & 0.66 & 0.55 & 0.65 & \maxval{0.68} & 0.24 & 0.28 & 0.20 & 0.27 & \maxval{0.33} \\
South Africa  & 0.80 & 0.80 & 0.78 & 0.81 & \maxval{0.82} & 0.53 & \maxval{0.56} & 0.50 & \maxval{0.56} & 0.54 \\
Spain         & 0.73 & 0.70 & 0.69 & 0.75 & \maxval{0.83} & 0.24 & 0.26 & 0.21 & 0.26 & \maxval{0.33} \\
Sweden        & 0.81 & 0.82 & 0.80 & 0.84 & \maxval{0.85} & 0.45 & 0.51 & 0.44 & 0.50 & \maxval{0.55} \\
Vietnam       & 0.46 & 0.30 & 0.31 & 0.46 & \maxval{0.67} & 0.15 & 0.09 & 0.08 & 0.15 & \maxval{0.22} \\
\bottomrule
\end{tabular}%
}
\label{tab:country_level_supp}
\end{table*}

\section{Mosaicking and Large Scale Inference}
\label{supp:mosaicking_large_scale_inference}

Deploying PRUE at the country scale requires constructing spatially complete, cloud-free Sentinel-2 composites from irregularly sampled, partially cloudy observations.
This section details our operational pipeline for scene selection, temporal compositing, and imagery quality mosaicking
using latitude-based season heuristics, greedy scene selection to minimize redundancy, and cloud-optimized data 
formats enabling scalable parallel inference.

\paragraph{Latitude-Based Season Heuristics.}

Planting and harvest windows were estimated using latitude-dependent day-of-year (DOY) ranges that account for hemispheric differences and climatic zones. This heuristic approach provides reasonable temporal constraints for scene selection, although we acknowledge that date selection could be substantially improved by integrating additional information on geographical variation in crop growth cycles, such as crop calendars, phenological models, or ground-based information on local planting and harvest periods.

\begin{figure}[H]
\centering
\captionof{algorithm}{Planting Season Heuristic Algorithm}
\resizebox{0.9\linewidth}{!}{%
\begin{minipage}{\linewidth}
{\footnotesize
\begin{algorithm}[H]
\begin{algorithmic}
\Function{PlantingSeasonDOY}{latitude}
    \State $\text{abs\_lat} \gets |\text{latitude}|$
    \If{$\text{abs\_lat} > 45$} \cmt{High latitudes}
        \State \Return $(91, 151)$ if $\text{latitude} > 0$ else $(274, 334)$
    \ElsIf{$20 < \text{abs\_lat} \leq 45$} \cmt{Mid-latitudes}
        \State \Return $(60, 120)$ if $\text{latitude} > 0$ else $(244, 334)$
    \ElsIf{$5 < \text{abs\_lat} \leq 20$} \cmt{Subtropics}
        \State \Return $(121, 212)$ if $\text{latitude} > 0$ else $(305, 365)$
    \Else \cmt{Equatorial $|\text{lat}| \leq 5$}
        \State \Return $(60, 121)$
    \EndIf
\EndFunction
\end{algorithmic}
\end{algorithm}
}
\end{minipage}%
}
\end{figure}

\begin{figure}[H]
\centering
\captionof{algorithm}{Harvest Season Heuristic Algorithm}
\resizebox{0.9\linewidth}{!}{%
\begin{minipage}{\linewidth}
{\footnotesize
\begin{algorithm}[H]
\begin{algorithmic}
\Function{HarvestSeasonDOY}{latitude}
    \State $\text{abs\_lat} \gets |\text{latitude}|$
    \If{$\text{abs\_lat} > 45$} \cmt{High latitudes}
        \State \Return $(244, 304)$ if $\text{latitude} > 0$ else $(60, 151)$
    \ElsIf{$20 < \text{abs\_lat} \leq 45$} \cmt{Mid-latitudes}
        \State \Return $(213, 304)$ if $\text{latitude} > 0$ else $(32, 120)$
    \ElsIf{$5 < \text{abs\_lat} \leq 20$} \cmt{Subtropics}
        \State \Return $(274, 365)$ if $\text{latitude} > 0$ else $(91, 181)$
    \Else \cmt{Equatorial $|\text{lat}| \leq 5$}
        \State \Return $(182, 243)$
    \EndIf
\EndFunction
\end{algorithmic}
\end{algorithm}
}
\end{minipage}%
}
\end{figure}

\paragraph{Feature Selection via Greedy Search}

Optimal scene selection maximizes spatial coverage while minimizing redundancy. Input scenes were pre-filtered by cloud cover ($<75\%$) and Scene Classification Layer (SCL) quality flags (excluding classes 1, 3, 7, 8, 9, 10, and nodata=0). The greedy approach prioritizes scenes that contribute valid observations to underrepresented spatial regions within a tile, enabling the use of relaxed scene-level cloud cover thresholds. Scenes with high overall cloud cover may still contain substantial cloud-free areas that fill critical gaps in the composite, thereby improving spatial completeness without requiring additional acquisitions.

\paragraph{Cloud-Optimized GeoTIFF Storage.}

For each Sentinel-2 grid tile and temporal period, median composites were constructed from the selected scenes. Spectral bands (B02, B03, B04, B08) at native 10 m resolution were masked using SCL upsampled from 20 m via nearest-neighbor. Temporal medians were computed alongside valid observation counts. Outputs were stored as float32 Cloud-Optimized GeoTIFFs with $1024 \times 1024$ internal tiling.

\paragraph{GTI-Based Reprojection, Resampling and Zarr Assembly.}

GDAL Tile Index (GTI)\footnote{\url{https://gdal.org/en/latest/drivers/raster/gti.html}} files provide virtual mosaics referencing distributed COGs. During Zarr\footnote{\url{https://zarr.readthedocs.io/en/stable/}} construction,

\begin{figure}[H]
\centering
\captionof{algorithm}{Greedy Scene Selection Algorithm}
\resizebox{0.9\linewidth}{!}{%
\begin{minipage}{\linewidth}
{\footnotesize
\begin{algorithm}[H]
\begin{algorithmic}
\Function{SelectScenesGreedy}{$\text{valid\_mask}$, $\text{target\_coverage}=5$, $\text{max\_scenes}=10$}
    \State \textit{Input:} $\text{valid\_mask}$ - boolean array of shape $(T, H, W)$
    \State $\text{coverage\_depth} \gets \mathbf{0}_{H \times W}$
    \State $\text{remaining} \gets \{0, 1, \ldots, T-1\}$
    \State $\text{selected} \gets []$
    \For{$i = 1 \rightarrow \text{max\_scenes}$}
        \State $\text{best\_idx} \gets \textsc{null}$
        \State $\text{best\_gain} \gets -1$
        \For{\textbf{each} $\text{idx} \in \text{remaining}$}
            \State $\text{undercovered} \gets (\text{coverage\_depth} < \text{target\_coverage})$
            \State $\text{new\_valid} \gets \text{valid\_mask}[\text{idx}] \land \text{undercovered}$
            \State $\text{gain} \gets \sum \text{new\_valid}$
            \If{$\text{gain} > \text{best\_gain}$}
                \State $\text{best\_gain} \gets \text{gain}$
                \State $\text{best\_idx} \gets \text{idx}$
            \EndIf
        \EndFor
        \If{$\text{best\_gain} = 0$}
            \State \textbf{break}
        \EndIf
        \State $\text{selected}.\textsc{append}(\text{time}[\text{best\_idx}])$
        \State $\text{coverage\_depth} \gets \text{coverage\_depth} + \text{valid\_mask}[\text{best\_idx}]$
        \State $\text{remaining}.\textsc{remove}(\text{best\_idx})$
    \EndFor
    \State \Return $\text{selected}$
\EndFunction
\end{algorithmic}
\end{algorithm}
}
\end{minipage}%
}
\end{figure}

\noindent\texttt{gdal\_translate} performs windowed extraction with on-the-fly reprojection to EPSG:3857 (Web Mercator) using nearest neighbor resampling and writes to a temporary local file. 
Reprojection to EPSG:3857 prior to inference eliminates the need for downstream pipelines to perform coordinate transformations, and enables global non-overlapping results without tile artifacts. The windowed data is then loaded and inserted directly into the Zarr store, with spatial partitions written to Zarr v3 arrays in parallel with Ray\footnote{\url{https://docs.ray.io/en/latest/index.html}}. This creates a robust and scalable approach to building large-scale, cloud-optimized data ready for downstream analysis.

\section{Large Scale Inference Visual Samples}
\label{supp:qualitative_examples} 

\begin{figure*}[htpb!]
  \centering
  \includegraphics[width=0.85\linewidth]{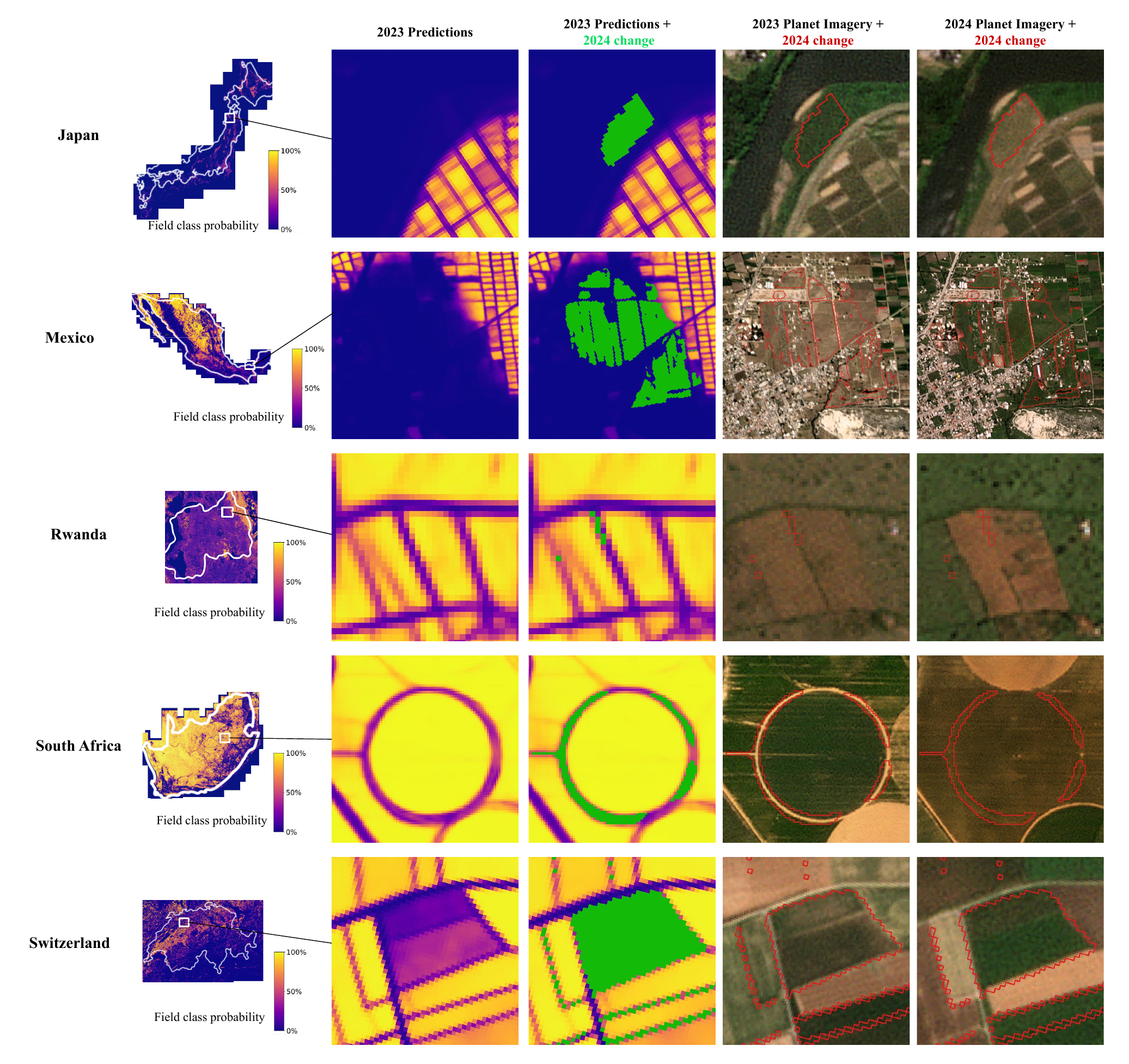}
  \caption{Example visuals over each country in our large-scale inference set (Japan, Mexico, Rwanda, South Africa, and Switzerland). For each  region of interest, we show: (1) the PRUE field boundary predictions from 2023; (2) the 2023 predictions with changes detected in 2024 highlighted in bright green; (3) Planet monthly basemaps from 2023 with the vectorized change mask outlined in red; and (4) Planet monthly basemaps from 2024 with the vectorized change mask outlined in red. The basemaps shown for each pair are from the same month in consecutive years.}
  \label{fig:countr_inference_examples}
\end{figure*}

To qualitatively assess PRUE's performance at operational scale, we present in Figure \ref{fig:countr_inference_examples} representative visual samples from each country-scale deployment described in Section~\ref{sec:ai_generated_boundaries}. These samples illustrate model behavior across diverse agricultural systems (Japan, Mexico, Rwanda, South Africa, Switzerland) spanning a wide range of field sizes. The visualizations demonstrate PRUE's ability to maintain spatial consistency across large extents under atmospheric variation. Visual inspection reveals spatial patterns of success and failure modes that inform future improvements.

\section{Change Detection Analysis}
\label{supp:change_detection}

The change detection visualizations presented in Figure~\ref{fig:countr_inference_examples} are intended to demonstrate how multi-year maps produced by PRUE can signal probable field-scale changes. We leave more detailed studies of change detection to future work. 

The method computes the absolute difference between semantic logits from consecutive years, applies min--max normalization, and thresholds at 0.5 to obtain a binary change mask. For well-calibrated models, this threshold highlights high-confidence semantic shifts. Visual inspection confirms that even small-scale detected changes are consistent with cultivation shifts (e.g., fields appearing or disappearing between years), and that artifacts from misregistration and atmospheric variation are uncommon. We note that lacking ground truth change labels, we relied on photo-interpretation of high-resolution basemap imagery rather than quantitative accuracy assessments, which we leave as future work.

\section{Future Directions}
\label{supp:future_directions}
Several directions remain open for future work, including: (1) comprehensive object-level~\cite{radoux2017good} and thematic accuracy~\cite{olofsson2014good} assessments on country-scale deployments with independent reference data; (2) exploring deployment metrics as out-of-distribution detectors (our preliminary findings suggest low consistency scores may signal when models encounter dissimilar data); (3) incorporating super-resolution approaches for delineating smallholder fields from temporal stacks of imagery; and (4) systematic post-processing ablations (morphological operations, topological cleaning, confidence thresholding) to further improve boundary quality in challenging regions.

\end{document}